\documentclass[10pt,twocolumn,letterpaper]{article}

\usepackage{cvpr}
\usepackage{times}
\usepackage{epsfig}
\usepackage{graphicx}
\usepackage{amsmath}
\usepackage{amssymb}
\usepackage{multirow}
\usepackage{pifont}
\usepackage{booktabs}
\usepackage{epsfig}
\usepackage{booktabs,multirow}
\usepackage{graphics}
\usepackage{threeparttable}
\usepackage{color}
\usepackage[normalem]{ulem}
\usepackage{multirow}
\usepackage{float}
\usepackage{amsfonts}
\usepackage{bm}
\usepackage{subfig}
\usepackage{enumitem}
\usepackage[numbers]{natbib}
\usepackage{array}
\usepackage[table]{xcolor}
\usepackage{colortbl}
\usepackage[export]{adjustbox}
\usepackage{bm}
\usepackage{amsmath}
\usepackage{gensymb}
\usepackage{wrapfig}
\usepackage{appendix}
\usepackage[noend]{algpseudocode}
\usepackage{algorithmicx,algorithm}
\newcolumntype{I}{!{\vrule width 1pt}}

\newcommand{\makesupptitle}[1]{
	\twocolumn[
	\begin{center}
		{\Large \bf #1 \par}
		{
		\large
		\lineskip .5em
		\par
		}
		\vskip .5em
		\vspace*{12pt}
	\end{center}
	]
}
\makeatletter
\newcommand{\thickhline}{%
    \noalign {\ifnum 0=`}\fi \hrule height 1pt
    \futurelet \reserved@a \@xhline
}
\makeatother
\newcommand{\tabincell}[2]{\begin{tabular}{@{}#1@{}}#2\end{tabular}}

\newcommand{\sssection}[1]{\noindent\textbf{#1}}
\newcommand{\pub}[1]{\color{gray}{\tiny{[{#1}]}}}
\newcommand{\cmark}{\ding{52}}%
\usepackage{caption}
\usepackage[pagebackref=true,breaklinks=true,letterpaper=true,colorlinks,bookmarks=false]{hyperref}

\usepackage[utf8]{inputenc}

\definecolor{bblue}{RGB}{0,30,95}
\definecolor{rred}{RGB}{190,0,0}
\definecolor{mygray}{gray}{.9}
\definecolor{ggray}{RGB}{127,127,127}

\usepackage{cleveref}
\crefname{section}{§}{§§}
\Crefname{section}{§}{§§}
\cvprfinalcopy 

\def\cvprPaperID{9564} 

\ifcvprfinal\fi
\begin{document}


\title{$_{\!\!\!\!\!\!}$\textsc{Lana}: A Language-Capable Navigator for Instruction Following and Generation$_{\!\!\!\!\!\!}$}
\author{{Xiaohan Wang
\quad Wenguan Wang
\quad Jiayi Shao
\quad Yi Yang}\thanks{Corresponding author: Yi Yang.} \\
\small{ReLER, CCAI, Zhejiang University} \\
\tt\small \url{https://github.com/wxh1996/LANA-VLN}
}

\maketitle

\begin{abstract}
Recently, visual-language navigation (VLN) -- entailing robot agents to follow navigation instructions -- has shown great$_{\!}$ advance.$_{\!}$ However,$_{\!}$ existing$_{\!}$ literature$_{\!}$ put$_{\!}$ most$_{\!}$ empha-
sis on interpreting instructions into actions, only delivering$_{\!}$ ``\textbf{dumb}''$_{\!}$ wayfinding$_{\!}$ agents.$_{\!}$ In$_{\!}$ this$_{\!}$ article,$_{\!}$ we$_{\!}$ devise$_{\!}$ \textbf{\textsc{Lana}}, a$_{\!}$ \textbf{\underline{la}nguage-capable}$_{\!}$ \underline{n}avigation$_{\!}$ \underline{a}gent$_{\!}$ which$_{\!}$ is$_{\!}$ able$_{\!}$ to$_{\!}$ not
only~execute human-written navigation commands, but also provide$_{\!}$ route$_{\!}$ descriptions$_{\!}$ to$_{\!}$ humans.$_{\!}$ This$_{\!}$ is$_{\!}$ achieved$_{\!}$ by$_{\!}$~si- multaneously$_{\!}$ learning$_{\!}$ instruction$_{\!}$ following$_{\!}$ and$_{\!}$ generation with$_{\!}$ only$_{\!}$ one$_{\!}$ \textbf{single}$_{\!}$ model.$_{\!}$ More$_{\!}$ specifically,$_{\!}$ two$_{\!}$ encoders, respectively$_{\!}$ for$_{\!}$~route$_{\!}$ and$_{\!}$ language$_{\!}$ encoding,$_{\!}$ are$_{\!}$ built$_{\!}$ and
shared$_{\!}$ by$_{\!}$ two$_{\!}$ decoders,$_{\!}$ respectively$_{\!}$ for$_{\!}$ action$_{\!}$ prediction
 and$_{\!}$  instruction$_{\!}$ generation,$_{\!}$ so$_{\!}$ as$_{\!}$ to$_{\!}$ exploit$_{\!}$ cross-task$_{\!}$ know-
 ledge$_{\!}$ and$_{\!}$ capture$_{\!}$ task-specific$_{\!}$ characteristics.$_{\!}$ Throughout pretraining and fine-tuning, both instruction following and generation$_{\!}$ are$_{\!}$ set$_{\!}$ as$_{\!}$ optimization$_{\!}$ objectives.$_{\!}$  We$_{\!}$ empirically
 verify that, compared with recent advanced task-specific solutions,$_{\!}$ {\textsc{Lana}}$_{\!}$ attains$_{\!}$ better$_{\!}$ performances on both instruction following and route description, with nearly half complexity. In addition, endowed with language generation capability,$_{\!}$ {\textsc{Lana}}$_{\!}$ can$_{\!}$ explain$_{\!}$ to$_{\!}$ human$_{\!}$ its$_{\!}$ behaviours$_{\!}$ and$_{\!}$ as- sist$_{\!}$~human's$_{\!}$ wayfinding.$_{\!}$ This$_{\!}$ work$_{\!}$ is$_{\!}$ expected$_{\!}$ to$_{\!}$ foster$_{\!}$ fu-
ture$_{\!}$ efforts$_{\!}$ towards$_{\!}$ building$_{\!}$ more$_{\!}$ trustworthy$_{\!}$ and$_{\!}$ socially-intelligent navigation robots.
\end{abstract}

	\vspace{-10pt}
\section{Introduction}

Developing agents that can interact with humans in natural language while perceiving and taking actions in their~environments is one of the fundamental goals in artificial intel- ligence.$_{\!}$ As$_{\!}$ a$_{\!}$ small$_{\!}$ step$_{\!}$ towards$_{\!}$ this$_{\!}$ target,$_{\!}$ visual-language navigation$_{\!}$ (VLN)$_{\!}$~\cite{anderson2018vision} -- endowing agents to execute natural language navigation commands -- recently received signifi- cant attention. In VLN space, much work has been done~on \textit{language grounding} -- teaching agents how to relate human instructions with actions associated with perceptions. How-
ever,$_{\!}$ there$_{\!}$ has$_{\!}$ been$_{\!}$ far$_{\!}$ little$_{\!}$ work$_{\!}$~\cite{fried2018speaker,tan2019learning,agarwal2019visual,wang2022counterfactual,dou2022foam} on$_{\!}$ the$_{\!}$ reverse$_{\!}$ side$_{\!}$ --$_{\!}$ \textit{language$_{\!}$ generation}$_{\!}$ --$_{\!}$ teaching$_{\!}$ agents how$_{\!}$ to verbalize$_{\!}$ a$_{\!}$ vivid$_{\!}$ description$_{\!}$ of$_{\!}$ navigation$_{\!}$ routes.$_{\!}$ More$_{\!}$~criti-
cally, existing VLN literature separately train agents that~are specialized$_{\!}$ for$_{\!}$ each$_{\!}$ single$_{\!}$ task.$_{\!}$ As$_{\!}$ a$_{\!}$ result,$_{\!}$ the$_{\!}$ delivered agents$_{\!}$ are either strong wayfinding actors but never talking, or conversable route instructors but never walking.

\begin{figure}[t]
	\vspace{-10pt}
	\begin{center}
		\includegraphics[width=\linewidth]{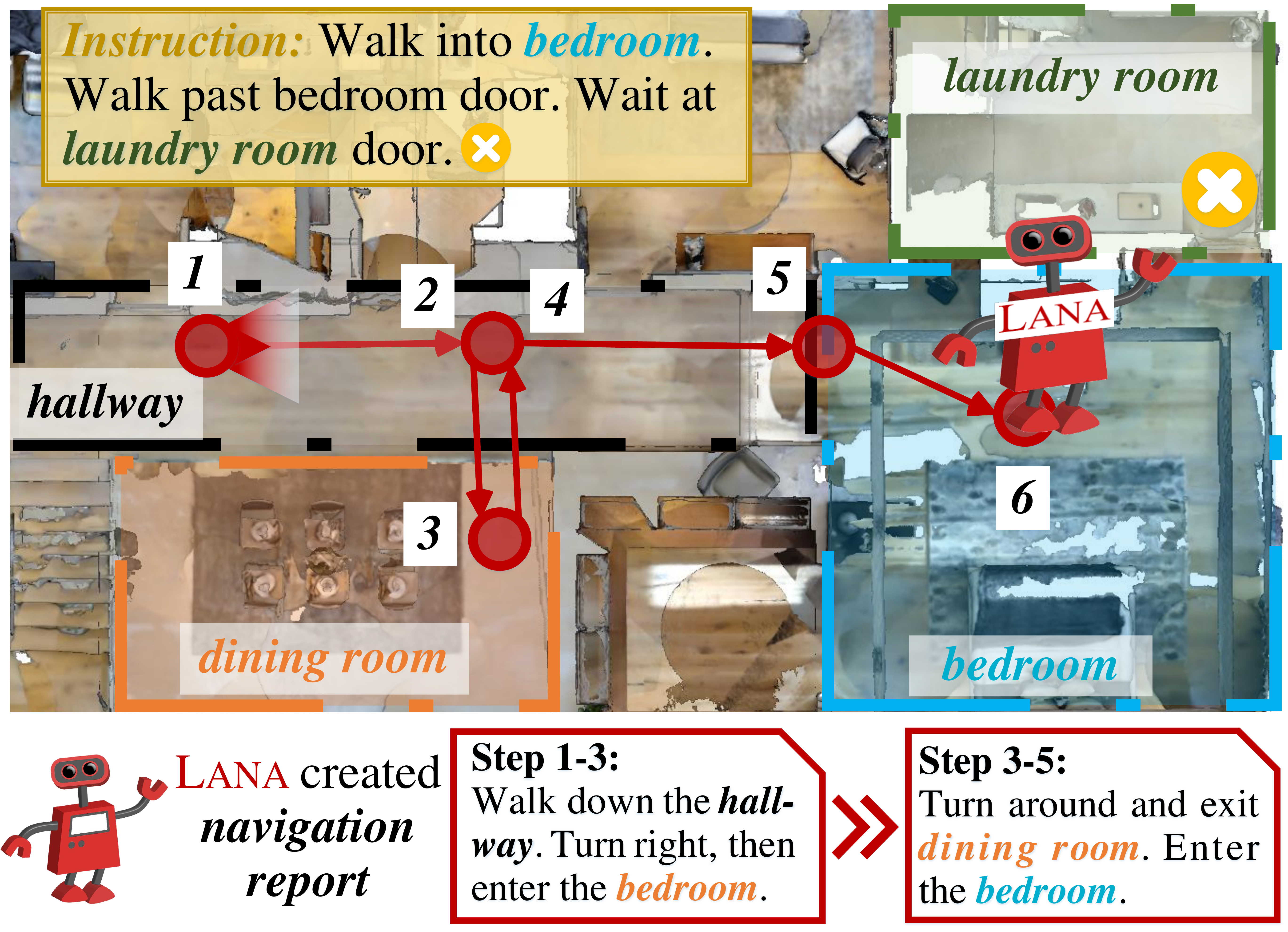}
	\end{center}
	\vspace{-18pt}
	\captionsetup{font=small}
	\caption{\small{\textsc{Lana} is capable of both instruction following and generation. Its written report benefits human-robot collaboration, and, to some extend,  can explain its behavior: it takes a wrong action at step 2 as it mistakes the dining room for bedroom. After gathering more information at step 3, it changes to the correct direction.}}
	\vspace{-12pt}
	\label{fig:fig1}
\end{figure}

This$_{\!}$ article$_{\!}$ underlines$_{\!}$ a$_{\!}$ fundamental$_{\!}$ challenge$_{\!}$ in$_{\!\!}$ VLN: \textit{Can we learn a single agent that is capable of both navigation instruction following and route description creation?}

We$_{\!}$ propose$_{\!}$ \textsc{Lana},$_{\!}$ a$_{\!}$ \underline{la}nguage-capable$_{\!}$ \underline{n}avigation$_{\!}$ \underline{a}gent, that$_{\!}$ is$_{\!}$ fully$_{\!}$ aware$_{\!}$ of$_{\!}$ such$_{\!}$ challenge$_{\!}$ (Fig.$_{\!\!}$~\ref{fig:fig1}).$_{\!}$ By$_{\!}$ simultane-
ously$_{\!}$ learning$_{\!}$ instruction$_{\!}$ grounding$_{\!}$ and$_{\!}$ generation,$_{\!}$ \textsc{Lana} formalises \textit{human-to-robot}$_{\!}$ and$_{\!}$ \textit{robot-to-human}$_{\!}$ communi- cation,$_{\!}$ conveyed$_{\!}$ using$_{\!}$ navigation-oriented$_{\!}$ natural$_{\!}$ language,$_{\!\!}$
 in$_{\!}$ a unified$_{\!}$ framework.$_{\!}$ This$_{\!}$ is$_{\!}$ of$_{\!}$ great$_{\!}$ importance,$_{\!}$ because:
\textbf{i)}$_{\!}$~It completes$_{\!}$ the$_{\!}$ necessary$_{\!}$ communication$_{\!}$ cycle$_{\!}$ between human$_{\!}$ and$_{\!}$ agents,$_{\!}$ and$_{\!}$  promotes$_{\!}$ VLN$_{\!}$ agent's real-world utility$_{\!}$~\cite{osswald2014learning}. For instance, when an agent takes long time to execute a navigation command, during which sustained human attention is infeasible and undesirable, the agent should report$_{\!}$ its$_{\!}$ progress$_{\!}$~\cite{thomason2020vision}.$_{\!}$ Also,$_{\!}$ agents$_{\!}$ are$_{\!}$ expected$_{\!}$ to$_{\!}$ direct human$_{\!}$ in$_{\!}$ agents'$_{\!}$ explored$_{\!}$ areas$_{\!}$~\cite{wang2022less}, which is relevant for search and rescue robots in  disaster regions$_{\!}$~\cite{tellex2011understanding,daniele2017navigational}, guide robots in public spaces$_{\!}$~\cite{wang2022counterfactual}, and navigation devices for the visually impaired$_{\!}$~\cite{huang2022assister}. \textbf{ii)} Two-way communication is integral to tight
human-robot coordination (\ie, ``\textit{I will continue this way} $\cdots$")~\cite{arkin2020multimodal}, and boosts human trust in robot~\cite{andrist2013rhetorical,dzindolet2003role}, hence$_{\!}$ increasing$_{\!}$ the$_{\!}$ acceptance$_{\!}$ of$_{\!}$ navigation$_{\!}$ robots.$_{\!}$ 
\textbf{iii)}$_{\!}$ Developing the language generation skill makes for more explainable robots, which can interpret their navigation beha- viors in a form of human-readable route descriptions.

$_{\!}$Technically,$_{\!}$ \textsc{Lana}$_{\!}$ is$_{\!}$ built$_{\!}$ as$_{\!}$ a$_{\!}$ Transformer-based,$_{\!}$ multi-task$_{\!}$ learning$_{\!}$ framework.$_{\!}$ The$_{\!}$ network$_{\!}$ consists$_{\!}$ of$_{\!}$
two$_{\!}$ uni-modal$_{\!}$ \textit{encoders}$_{\!}$ respectively$_{\!}$ for$_{\!}$ language$_{\!}$ and$_{\!}$ route$_{\!}$ encod- ing,$_{\!}$ and$_{\!}$ two$_{\!}$ multimodal$_{\!}$ \textit{decoders}$_{\!}$  respectively$_{\!}$ for$_{\!}$ route-to-instruction and instruction-to-route translation, based on the$_{\!}$ two$_{\!}$ encoders.$_{\!}$ The$_{\!}$ whole$_{\!}$ network$_{\!}$ is$_{\!}$ end-to-end$_{\!}$ learned with
the tasks of both instruction grounding and generation, during both pretraining and fine-tuning phases. Taken all these
together,$_{\!}$ \textsc{Lana}$_{\!}$ provides$_{\!}$ a$_{\!}$ unified,$_{\!}$ powerful$_{\!}$ framework$_{\!}$ that explores both$_{\!}$ task-specific$_{\!}$ and$_{\!}$ cross-task$_{\!}$ knowledge$_{\!}$ at$_{\!}$ the heart of model design and network training. \textsc{Lana}$_{\!}$ thus$_{\!}$ can better$_{\!}$ comprehend$_{\!}$ linguistic$_{\!}$ cues$_{\!}$ (\eg,$_{\!}$ words,$_{\!}$ phrases,$_{\!}$ and sentences), visual perceptions, actions over long temporal
 horizons and their relationships, even in the absence of ex- plicit$_{\!}$ supervision,$_{\!}$ and$_{\!}$ eventually$_{\!}$ benefits$_{\!}$ both$_{\!}$ the$_{\!}$ two$_{\!}$ tasks.

We conduct extensive experiments on three famous VLN datasets (\ie, R2R$_{\!}$~\cite{anderson2018vision}, R4R$_{\!}$~\cite{jain2019stay}, REVERIE$_{\!}$~\cite{2020REVERIE}), for both instruction$_{\!}$ following$_{\!}$ and$_{\!}$ generation,$_{\!}$ giving$_{\!}$ a$_{\!}$ few$_{\!}$ intriguing points:$_{\!}$ \textbf{First},$_{\!}$ \textsc{Lana}$_{\!}$ successfully$_{\!}$ solves$_{\!}$ the$_{\!}$ two$_{\!}$ tasks$_{\!}$ using  only one single agent, {without} switching between different
models.$_{\!}$ \textbf{Second},$_{\!}$ with$_{\!}$ an$_{\!}$ elegant$_{\!}$ and$_{\!}$ integrated$_{\!}$ architecture,
\textsc{Lana}$_{\!}$ performs$_{\!}$ comparable,$_{\!}$ or even$_{\!}$ better$_{\!}$ than$_{\!}$ recent$_{\!}$ top-leading,$_{\!}$ task-specific$_{\!}$ alternatives.$_{\!}$  \textbf{Third},$_{\!}$ compared$_{\!}$ to$_{\!}$ learn-
ing each task individually, training \textsc{Lana} on the two tasks jointly$_{\!}$ obtains$_{\!}$ better$_{\!}$ performance$_{\!}$ with$_{\!}$ reduced$_{\!}$ complexity and model size, confirming the advantage
of \textsc{Lana} in cross-task$_{\!}$ relatedness$_{\!}$ modeling$_{\!}$ and$_{\!}$ parameter$_{\!}$ efficiency.$_{\!}$ \textbf{Forth}, \textsc{Lana}$_{\!}$ can$_{\!}$ explain$_{\!}$ to$_{\!}$ human$_{\!}$ its$_{\!}$ behavior$_{\!}$ by$_{\!}$ verbally$_{\!}$ describing its navigation routes. \textsc{Lana} can be essentially viewed as an explainable VLN robot, equipped with a  self-adaptively trained language explainer. \textbf{Fifth}, subjective analyses reveal our linguistic outputs are of higher quality than the baselines but still  lag behind human-generated$_{\!}$ utterances.$_{\!}$ While$_{\!}$ there
 is$_{\!}$ still$_{\!}$ room$_{\!}$ for$_{\!}$ improvement,$_{\!}$ our$_{\!}$ results$_{\!}$ shed$_{\!}$ light$_{\!}$ on$_{\!}$ a$_{\!}$ pro- mising$_{\!}$ direction$_{\!}$ of$_{\!}$ future$_{\!}$ VLN$_{\!}$ research,$_{\!}$ with$_{\!}$ great$_{\!}$ poten- tial$_{\!}$ for$_{\!}$ explainable$_{\!}$ navigation$_{\!}$ agents$_{\!}$ and$_{\!}$ robot$_{\!}$ applications.$_{\!}$

\section{Related Work}
\noindent\textbf{Navigation Instruction Following.} Building autonomous, language-based navigation agents is a long-standing target for natural language processing and robotics communities. Rather than previous studies bounded to controlled~environ- mental$_{\!}$ context$_{\!}$~\cite{macmahon2006walk,tellex2011understanding,chen2011learning,andreas2015alignment,mei2016listen},$_{\!}$
Anderson$_{\!}$~\etal~$_{\!}$\cite{anderson2018vision}$_{\!}$ lift such$_{\!}$~task$_{\!}$ to$_{\!}$ a$_{\!}$ photo-realistic$_{\!}$ setting$_{\!}$ --$_{\!}$ VLN,$_{\!}$ stimulating$_{\!}$ in- creasing$_{\!}$ interest$_{\!}$ in$_{\!}$ computer$_{\!}$ vision$_{\!}$ field.$_{\!}$ Early$_{\!}$ efforts$_{\!}$ were built$_{\!}$ upon$_{\!}$ recurrent$_{\!}$ neural$_{\!}$ networks.$_{\!}$ They$_{\!}$ explore$_{\!}$ diverse$_{\!}$ training$_{\!}$ strategies$_{\!}$~\cite{wang2018look,wang2019reinforced},$_{\!}$ mine$_{\!}$ extra$_{\!}$ supervisory$_{\!}$ signals from synthesized samples$_{\!}$~\cite{fried2018speaker,tan2019learning,fu2019counterfactual} or auxiliary tasks \cite{wang2019reinforced,huang2019transferable,ma2019self,zhu2019vision,wang2022counterfactual},$_{\!}$ and$_{\!}$ explore$_{\!}$ intelligent$_{\!}$ path$_{\!}$ planning$_{\!}$ \cite{ke2019tactical,ma2019regretful,wang2020active}.$_{\!}$ For$_{\!}$ structured$_{\!}$ and$_{\!}$ long-range$_{\!}$ context$_{\!}$ modeling, recent$_{\!}$
solutions$_{\!}$ were$_{\!}$ developed$_{\!}$ with$_{\!}$ environment map~\cite{zhao2022target,chen2022think,deng2020evolving,wang2021structured},$_{\!}$ transformer$_{\!}$ architectures$_{\!}$~\cite{hong2020recurrent,pashevich2021episodic,lin2022multimodal,qiao2022hop,chen2021history}, and  multi-modal pretraining \cite{majumdar2020improving,hao2020towards,guhur2021airbert,chen2022learning}.

Unlike existing VLN solutions that are \textit{all} specialized for navigation instruction following, we are ambitious to build~a powerful agent that is able to not only execute navigation~in- structions but also describe its navigation routes. We stick~to
 this target throughout our algorithm -- from network design, to model pretraining, to fine-tuning. Through jointly learning instruction execution and generation, our agent can bet- ter$_{\!}$ ground$_{\!}$ instructions$_{\!}$ into$_{\!}$ perception$_{\!}$ and$_{\!}$ action,$_{\!}$ and,$_{\!}$ to$_{\!}$ certain degree, interpret its behavior and foster human trust. Our target and visual-dialog navigation~\cite{thomason2020vision} are different (yet complementary), as the latter only focuses on the situation where agents use language to ask for human assistance.

\noindent\textbf{Navigation$_{\!}$ Instruction$_{\!}$ Generation.$_{\!}$} The$_{\!}$ study$_{\!}$ of$_{\!}$ instruc- tion$_{\!}$ creation$_{\!}$~\cite{curry2015generating}$_{\!}$ can$_{\!}$ date$_{\!}$ back$_{\!}$ to$_{\!}$ the$_{\!}$ 1960s$_{\!}$~\cite{Lynch}.$_{\!}$ Early$_{\!}$~work \cite{ward1986turn,allen1997knowledge,lovelace1999elements}$_{\!}$ found$_{\!}$ human$_{\!}$ route$_{\!}$ direction$_{\!}$ is$_{\!}$ tied$_{\!}$ to$_{\!}$ cognitive$_{\!}$ map$_{\!}$~\cite{kuipers1978modeling}, and impacted$_{\!}$ by$_{\!}$ many$_{\!}$ factors,$_{\!}$ \eg,$_{\!}$ cultural$_{\!}$ back- ground$_{\!}$~\cite{vanetti1988communicating}$_{\!}$ and$_{\!}$ gender$_{\!}$~\cite{hund2006getting}.$_{\!}$ They$_{\!}$ also$_{\!}$ reached$_{\!}$ a$_{\!}$ consensus
that involving \textit{turn-by-turn directions} and \textit{salient landmarks}  makes instructions easier for human to follow \cite{look2005location,waller2007landmarks,richter2008simplest}. Based on these efforts, a few computational systems are developed using pre-built \textit{templates}$_{\!}$~\cite{look2005location,goeddel2012dart}, or hand-crafted \textit{rules}$_{\!}$~\cite{dale2004using}. Though providing high-quality output in targeted scenarios, they$_{\!}$ require$_{\!}$ expertise$_{\!}$ of$_{\!}$ linguistic$_{\!}$ knowledge$_{\!}$ and extensive$_{\!}$ effort$_{\!}$ for$_{\!}$ building$_{\!}$ the$_{\!}$ templates/rules.$_{\!}$ Some$_{\!}$ data-driven solutions~\cite{cuayahuitl2010generating,osswald2014learning,daniele2017navigational,fried2017unified} emerged later, yet confined to$_{\!}$ simplified$_{\!}$ grid-like$_{\!}$ or$_{\!}$ perception-poor$_{\!}$ environments.

Generating$_{\!}$ natural$_{\!}$ language$_{\!}$ instructions$_{\!}$ has$_{\!}$ long$_{\!}$ been viewed$_{\!}$ as a core functionality of socially intelligent robots and$_{\!}$
been$_{\!}$ of$_{\!}$ great$_{\!}$ interest$_{\!}$ in$_{\!}$ many$_{\!}$ disciplines$_{\!}$ such$_{\!}$ as$_{\!}$ robo- tics$_{\!}$~\cite{goeddel2012dart},$_{\!}$ linguistics$_{\!}$~\cite{striegnitz2011report},$_{\!}$ cognition\!~\cite{kuipers1978modeling,evans1981environmental},$_{\!}$ psychology$_{\!}$~\cite{vanetti1988communicating},$_{\!}$ and$_{\!}$ geo$_{\!}$ science$_{\!}$~\cite{de2021identifying}. Surprisingly little has been done in the field$_{\!}$ of$_{\!}$ embodied$_{\!}$ vision.$_{\!}$ For$_{\!}$ the$_{\!}$ rare$_{\!}$ exceptions$_{\!}$~\cite{fried2018speaker,tan2019learning,schumann2021generating,agarwal2019visual,wang2022counterfactual,dou2022foam},$_{\!}$ \cite{fried2018speaker,tan2019learning,schumann2021generating}$_{\!}$ are$_{\!}$ only$_{\!}$ to$_{\!}$ augment$_{\!}$ the$_{\!}$ training$_{\!}$~data for$_{\!}$ boosting$_{\!}$ wayfinding,$_{\!}$ and,$_{\!}$ all$_{\!}$ of$_{\!}$ them$_{\!}$ learn$_{\!}$ a$_{\!}$ single agent specialized$_{\!}$ for$_{\!}$ instruction$_{\!}$ generation.$_{\!}$ Our$_{\!}$ idea$_{\!}$ is$_{\!}$ fundamen- tally$_{\!}$ different.$_{\!}$ We$_{\!}$ are$_{\!}$ to$_{\!}$ build$_{\!}$ a$_{\!}$ language-capable$_{\!}$ navigation agent$_{\!}$ that$_{\!}$ masters$_{\!}$ both$_{\!}$ instruction$_{\!}$ following$_{\!}$ and$_{\!}$ creation. As a result, this work represents an early yet solid attempt towards socially intelligent, embodied navigation robots.

\begin{figure*}[t]
	\vspace{-12pt}
	\begin{center}
		\includegraphics[width=\linewidth]{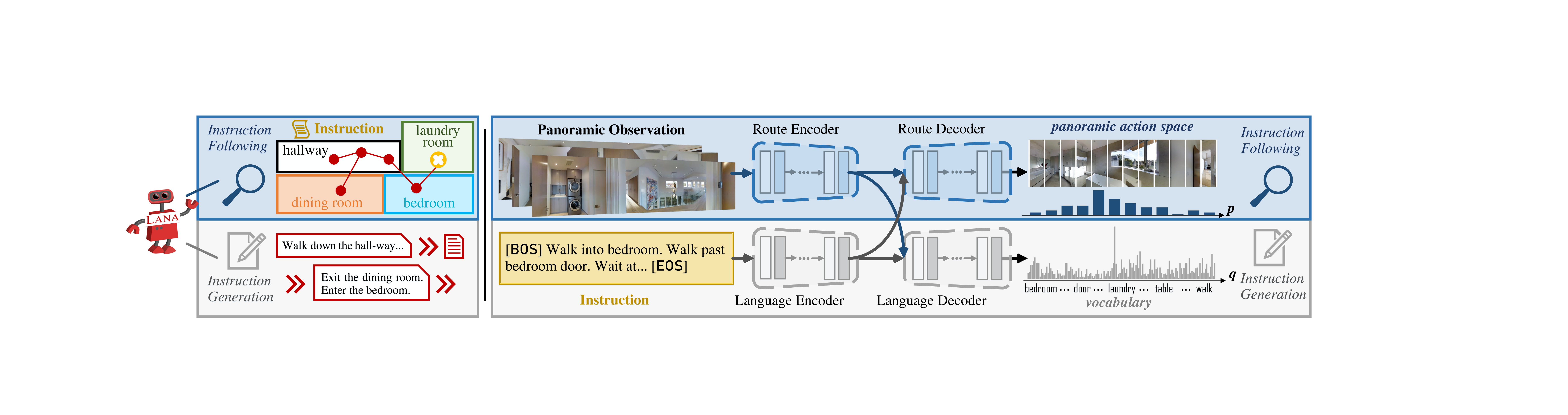}
\put(-197,78){\scriptsize $\mathcal{E}^r$}
\put(-195,6){\scriptsize $\mathcal{E}^l$}
\put(-137,78){\scriptsize $\mathcal{D}^r$}
\put(-136,6){\scriptsize $\mathcal{D}^l$}
	\end{center}
	\vspace{-18pt}
	\captionsetup{font=small}
	\caption{\small{Architecture overview of \textsc{Lana} (\S\ref{sec:ma}).}}
	\vspace{-16pt}
	\label{fig:fig2}
\end{figure*}
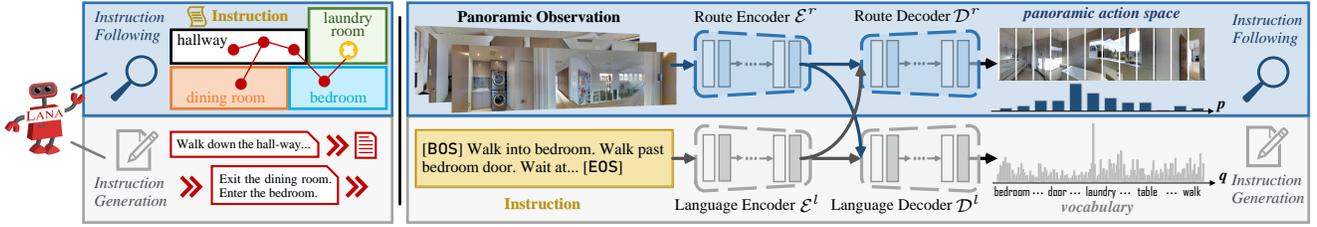

\noindent\textbf{Auxiliary$_{\!}$ Learning$_{\!}$ in$_{\!}$ VLN.$_{\!}$} There$_{\!}$ are$_{\!}$ several$_{\!}$ VLN$_{\!}$~solu- tions$_{\!}$~\cite{ma2019self,zhu2019vision,wangtowards}$_{\!}$ exploit$_{\!}$ extra$_{\!}$ supervision$_{\!}$ signals$_{\!}$ from$_{\!}$ auxi- liary$_{\!}$ tasks$_{\!}$ to$_{\!}$ aid$_{\!}$
navigation$_{\!}$ policy$_{\!}$ learning.$_{\!}$ For$_{\!}$ the$_{\!}$ auxiliary tasks,$_{\!}$ representative$_{\!}$ ones$_{\!}$ include$_{\!}$ next-step orientation reg- ression$_{\!}$~\cite{zhu2019vision},$_{\!}$ navigation$_{\!}$ progress$_{\!}$ estimation$_{\!}$~\cite{ma2019self},$_{\!}$ path$_{\!}$ back-translation$_{\!}$~\cite{zhu2019vision,wang2022counterfactual}, trajectory-instruction$_{\!}$ compatibility$_{\!}$ prediction \cite{zhu2019vision},  as well as final target localization$_{\!}$~\cite{zhao2022target}.

These$_{\!}$ VLN solutions put focus on instruction following; the$_{\!}$ auxiliary$_{\!}$ tasks$_{\!}$ are$_{\!}$ the$_{\!}$ means,$_{\!}$ not$_{\!}$ the$_{\!}$ end.$_{\!}$ By$_{\!}$ contrast, we~aim to build one single agent that learns to master both$_{\!}$ instruction$_{\!}$ following$_{\!}$ and$_{\!}$ creation$_{\!}$ well.$_{\!\!}$ Although$_{\!}$~\cite{wang2022counterfactual}$_{\!}$
pays equal attention to instruction following and generation under a dual-task learning scheme, it still learns two separate single-task agents. Moreover, the aforementioned auxiliary tasks, in principle, can be utilized by our agent.

\noindent\textbf{Vision-Language$_{\!}$ Pretraining$_{\!}$ for$_{\!}$ VLN.$_{\!}$} Vision-language$_{\!}$ pretraining~\cite{radford2021learning,tan2019lxmert,chen2020uniter} on massive-scale image-text pairs has$_{\!}$ recently$_{\!}$ witnessed$_{\!}$ rapid$_{\!}$ progress.$_{\!}$ It$_{\!}$ has$_{\!}$ been$_{\!}$ proven$_{\!}$~that transferable$_{\!}$ cross-modal$_{\!}$ representations$_{\!}$ can$_{\!}$ be$_{\!}$ delivered$_{\!}$~via such$_{\!}$ pretraining$_{\!}$ and$_{\!}$ facilitate$_{\!}$ downstream$_{\!}$ tasks~\cite{wang2021t2vlad,liang2022visual,zhao2022centerclip,radford2021learning,li2022blip,liang2022locater}.$_{\!}$ Such$_{\!}$ training$_{\!}$ regime$_{\!}$ has$_{\!}$ become$_{\!}$ increasingly$_{\!}$ popular$_{\!}$ in$_{\!}$ VLN.$_{\!}$ In$_{\!}$ particular,$_{\!}$ a$_{\!}$ few$_{\!}$ early$_{\!}$ endeavors$_{\!}$~\cite{li2019robust,hong2020recurrent} 
directly$_{\!}$ adopt general$_{\!}$ vision-language$_{\!}$ pretraining$_{\!}$ for$_{\!}$ VLN,$_{\!}$ without considering task-specific nature. Later, \cite{guhur2021airbert,hao2020towards,chen2022learning} conduct pretraining on abundant web image-captions~\cite{guhur2021airbert} or synthesized trajectory-instruction pairs \cite{hao2020towards,chen2022learning} with different VLN-specific proxy tasks. \cite{chen2021history,qiao2022hop} introduce history-aware proxy tasks for more VLN-aligned pretraining.

From the view of proxy task, existing VLN pretraining follows$_{\!}$ the$_{\!}$ \textit{masked$_{\!}$ language$_{\!}$ modeling}$_{\!}$ regime$_{\!}$~\cite{kenton2019bert}.$_{\!}$ Differ- ently,$_{\!}$ our$_{\!}$ pretraining$_{\!}$ is$_{\!}$ based$_{\!}$ on$_{\!}$ language$_{\!}$ generation,$_{\!}$ which helps$_{\!}$ the$_{\!}$ agent$_{\!}$ to$_{\!}$ capture$_{\!}$ the$_{\!}$ linguistic$_{\!}$ structures$_{\!}$ so$_{\!}$ as$_{\!}$ to$_{\!}$ reach$_{\!}$ comprehensive$_{\!}$ understanding$_{\!}$ of$_{\!}$ language$_{\!}$ commands and$_{\!}$ boost$_{\!}$ instruction$_{\!}$ execution.$_{\!}$ Recent$_{\!}$ {advance$_{\!}$~\cite{wang2021simvlm,yu2022coca,desai2021virtex,wang2022align}} in$_{\!}$ general$_{\!}$ vision-language$_{\!}$ pretraining$_{\!}$ also$_{\!}$ confirm$_{\!}$ the$_{\!}$ value of$_{\!}$ generative$_{\!}$ language$_{\!}$ modeling.$_{\!}$ Moreover,$_{\!}$ for$_{\!}$ \textsc{Lana},$_{\!}$ instruction$_{\!}$ generation is not merely a proxy task that is often dropped after pretraining, but also a main training target during fine-tuning, and the fundamental basis for the ability of language-based route direction during deployment.

\section{Methodology}
\noindent\textbf{Task$_{\!}$ Setup.$_{\!}$} Our target is to build a language-capable navi- gation agent, which masters$_{\!}$~both$_{\!}$ instruction$_{\!}$ following$_{\!}$ and generation$_{\!}$ tasks,$_{\!}$ using$_{\!}$ one single model instance only.
\begin{itemize}[leftmargin=*]
	\setlength{\itemsep}{0pt}
	\setlength{\parsep}{-2pt}
	\setlength{\parskip}{-0pt}
	\setlength{\leftmargin}{-10pt}
	\vspace{-5pt}
	\item \textit{Instruction following:$_{\!}$} the agent needs to find a route~$R\!=$ $\!\{r_{t}\}^T_{t=1\!\!}$ of$_{\!}$ $T_{\!}$ steps$_{\!}$ to$_{\!}$ a$_{\!}$ target$_{\!}$ location,$_{\!}$ following$_{\!}$ a$_{\!}$ human-written instruction $X\!=\!\{x_l\}^L_{l=1\!}$ of $L$ words. At~each~step $t$,$_{\!}$ the$_{\!}$ agent$_{\!}$ gets$_{\!}$ a$_{\!}$ panoramic$_{\!}$ RGB$_{\!}$ percept$_{\!}$ ${O}^t$,$_{\!}$ discretized into$_{\!}$ $K_{\!}\!=\!36_{\!}$ views,$_{\!}$ \ie,$_{\!}$ ${O}_{t\!}\!=\!\{o_{t,k}\!\in\!\mathbb{R}^{3\!\times\!224\!\times\!224\!}\}^K_{k=1}$.$_{\!}$~The agent$_{\!}$ selects$_{\!}$ a$_{\!}$ navigation$_{\!}$ action$_{\!}$ $a_t$,$_{\!}$ \ie,$_{\!}$ a$_{\!}$ navigable$_{\!}$ view,$_{\!}$ from$_{\!}$ $\!\{o_{t,k}\}_{k\!}$ to$_{\!}$ perform. Together, ${O}_t$ and $a_t$ determine the~route state $r_{t}$ at step $t$, \ie, $r_{t}\!=\!(O_{t}, a_t)$.
	\item \textit{Instruction$_{\!}$ generation:$_{\!}$} the$_{\!}$ agent$_{\!}$ observes$_{\!}$ a$_{\!}$ navigation route$_{\!}$ $R_{\!}\!=_{\!}\!\{r_{t}\}^T_{t=1}$,$_{\!}$ \ie,$_{\!}$ a$_{\!}$ sequence$_{\!}$ of$_{\!}$ actions$_{\!}$ $\{a_{t}\}^T_{t=1\!}$ along with the panoramic percept $\{O_{t}\}^T_{t=1}$, and must verbalize a grounded description $X\!=\!\{x_l\}^L_{l=1\!}$ of the route $R$.
	\vspace{-4pt}
\end{itemize}

\noindent\textbf{Method$_{\!}$ Overview.$_{\!}$} To$_{\!}$ address$_{\!}$ our$_{\!}$ challenging$_{\!}$ setting,$_{\!}$ we devise \textsc{Lana}, a Transformer-based, multi-task VLN agent (Fig.~\ref{fig:fig2}) that exploits cross-task commonalities throughout architecture design (\S\ref{sec:ma}) and network training (\S\ref{sec:nt}).

	\vspace{-2pt}
\subsection{Model Architecture}\label{sec:ma}
	\vspace{-1pt}
$_{\!}$\textsc{Lana}$_{\!}$ can$_{\!}$ take$_{\!}$ the$_{\!}$ navigation$_{\!}$ route$_{\!}$ $R_{\!}\!=_{\!}\!\{r_{t}\}_{t\!}$ as$_{\!}$ input$_{\!}$~and output$_{\!}$ corresponding$_{\!}$ description$_{\!}$ $X_{\!}\!=_{\!}\!\{x_l\}_{l}$,$_{\!}$ and$_{\!}$ vice$_{\!}$ versa. To$_{\!}$ achieve$_{\!}$ such$_{\!}$ \textit{bi-directional$_{\!}$ translation}$_{\!}$ between$_{\!}$ navigation route$_{\!}$ $R_{\!}$ and$_{\!}$ instruction$_{\!}$ $X$, and explore cross-task rela- tedness, \textsc{Lana} is elaborately designed as a composition of:
\begin{itemize}[leftmargin=*]
	\setlength{\itemsep}{0pt}
	\setlength{\parsep}{-2pt}
	\setlength{\parskip}{-0pt}
	\setlength{\leftmargin}{-10pt}
	\vspace{-6pt}
	\item \textit{Route$_{\!}$ Encoder$_{\!}$ $\mathcal{E}^{r\!}$} and$_{\!}$ \textit{Language$_{\!}$ Encoder$_{\!}$} $\mathcal{E}^{l\!}$ for$_{\!}$ unimoda- lity representation encoding based on self-attention; and
	\item \textit{Language$_{\!}$ Decoder}$_{\!}$ $\mathcal{D}^{l\!}$ and$_{\!}$ \textit{Route$_{\!}$ Decoder}$_{\!}$ $\mathcal{D}^{r\!}$ for$_{\!}$ cross-attention$_{\!}$ based$_{\!}$ route-language$_{\!}$ bi-directional$_{\!}$ translation.
	\vspace{-6pt}
\end{itemize}
The$_{\!}$ two$_{\!}$ unimodal$_{\!}$ encoders$_{\!}$ are$_{\!}$ shared$_{\!}$ between$_{\!}$ and$_{\!}$ jointly trained$_{\!}$ with$_{\!}$ the$_{\!}$ two$_{\!}$ multimodal$_{\!}$ decoders.$_{\!}$ They$_{\!}$ work$_{\!}$ closely:
\begin{itemize}[leftmargin=*]
	\setlength{\itemsep}{0pt}
	\setlength{\parsep}{-2pt}
	\setlength{\parskip}{-0pt}
	\setlength{\leftmargin}{-10pt}
	\vspace{-6pt}
	\item \textit{Instruction$_{\!}$ Following:}$_{\!}$ at$_{\!}$ navigation$_{\!}$ step$_{\!}$ $t$,$_{\!}$ \textsc{Lana}$_{\!}$ respectively feeds$_{\!}$ the$_{\!}$ entire$_{\!}$ instruction$_{\!}$ $X_{\!}$ and$_{\!}$~the$_{\!}$ sequence$_{\!}$ of historical$_{\!}$ route$_{\!}$ states$_{\!}$ $\{r_1,\cdots_{\!}, r_{t-1}\}_{\!}$ and current$_{\!}$ percept $O_{t}$ into$_{\!}$ the$_{\!}$ corresponding$_{\!}$ encoders,$_{\!}$ and$_{\!}$ utilizes$_{\!}$ the$_{\!}$ route$_{\!}$ decoder $\mathcal{D}^{r\!}$ to predict the navigation action $a_t$.
	\item \textit{Instruction$_{\!}$ Generation:}$_{\!}$ at$_{\!}$ generation$_{\!}$ step$_{\!}$ $l$,  \textsc{Lana}$_{\!}$ respec- tively$_{\!}$ feeds$_{\!}$ the$_{\!}$ full$_{\!}$ route$_{\!}$ $R_{\!}$ and$_{\!}$ prior$_{\!}$ predicted$_{\!}$ words$_{\!}$ $\{x_1,\cdots_{\!}, x_{l-1}\}_{\!}$ into$_{\!}$ the$_{\!}$ corresponding$_{\!}$ encoders,$_{\!}$ and$_{\!}$ uti- lizes the language decoder $\mathcal{D}^{l\!}$ to predict the next word $x_l$.
	\vspace{-17pt}
\end{itemize}
We remark that \textsc{Lana} conducts the instruction generation task in an \textit{autoregressive} manner. In this way,  both the two decoders are conditioned on both the two encoders, leading to extensive visual and linguistic knowledge exchange.

\noindent\textbf{Route$_{\!}$ Encoder$_{\!}$ $\mathcal{E}^{r\!}$.}$_{\!}$ $\mathcal{E}^{r\!}$ takes$_{\!}$ input$_{\!}$ either$_{\!}$ the$_{\!}$ entire$_{\!}$ route,$_{\!}$ \ie,$_{\!}$  {$R\!=\!\{r_{t}\}^T_{t=1\!}\!=\!\{(O_{t}, a_t)\}^T_{t=1}$,$_{\!}$ }during$_{\!}$ instruction$_{\!}$ generation; or historical$_{\!}$ route$_{\!}$ states$_{\!}$ along with current observation, \ie, $\{r_1,\cdots_{\!}, r_{t-1}, O_{t}\}_{\!}\!=_{\!}\!\{O_{1}, a_1,\cdots_{\!}, O_{t-1}, a_{t-1}, O_{t}\}$, during wayfinding. Hence it has two types of input tokens corresponding to the panoramic observation $O_{t}$ and action $a_{t}$. In particular, the observation token of $O_{t}$ is calculated by:
	\vspace{-5pt}
\begin{equation}\small
    \begin{aligned}\label{eq:1}
    \bm{O}_{t} &= [\bm{o}_{t,1}, \bm{o}_{t,2}, \cdots, \bm{o}_{t,K}]~~\in\mathbb{R}^{K\times d},\\
    \bm{o}_{t,k} &= \mathcal{F}^v(\bm{v}_{t,k}) + \mathcal{F}^\theta(\bm{\theta}_{t,k}) + \bm{\tau}^t + \bm{\tau}^O~~\in\mathbb{R}^{d},
    \end{aligned}
    	\vspace{-4pt}
\end{equation}
where$_{\!}$ $\bm{v}_{t,k\!}$ and$_{\!}$ $\bm{\theta}_{t,k\!}$ are$_{\!}$ respectively$_{\!}$ the$_{\!}$ visual$_{\!}$ and$_{\!}$ orienta- tion$_{\!}$ embeddings of view ${o}_{t,k}$; $\mathcal{F}^{v/\theta\!}$ is linear projection for feature dimension alignment; $\bm{\tau}^{t\!}\!\in\!\mathbb{R}^{d}$ embeds the temporal order -- $t$; and $\bm{\tau}^{O\!}\!\in\!\mathbb{R}^{d}$ is a learnable type embedding which indicates $\bm{o}_{t,k}$ is an observation token.

Similarly, the action token of $a_{t}$ is given as:
	\vspace{-4pt}
\begin{equation}\small
    \begin{aligned}\label{eq:2}
    \bm{a}_{t} = \mathcal{F}^v(\bm{v}_{t,a_{t}}) + \mathcal{F}^\theta(\bm{\theta}_{t,a_{t}}) + \bm{\tau}^t + \bm{\tau}^A~~\in\mathbb{R}^{d},
    \end{aligned}
    \vspace{-3pt}
\end{equation}
where $\bm{v}_{t,a_{t}\!}$ and $\bm{\theta}_{t,a_{t}\!}$ respectively embed the visual view and turned angle that are associated with action $a_{t}$. Analogous to Eq.~\ref{eq:1}, $\bm{\tau}^A\!\in\!\mathbb{R}^{d}$ encodes the action token type.

Tokenizing all the $K_{\!}$ subviews $\{\bm{o}_{t,k}\}_{k\!}$ of each panoramic percept $O_{t}$ allows \textsc{Lana} to access/memorize all the observations along the navigation route. Unfortunately, consider- ing$_{\!}$ such$_{\!}$ many$_{\!}$ tokens$_{\!}$ causes$_{\!}$ unaffordable$_{\!}$ computation$_{\!}$ load for self-attention based encoding. To pursue a good balance between computational cost and representation ability, we first compute an action-attentive route state:
 \vspace{-4pt}
\begin{equation}\small
    \begin{aligned}\label{eq:3}
    \!\!\!\bm{r}_{t} &\!=\!\bm{a}_{t} + \bm{c}_{t}~\in\mathbb{R}^{d},\\
    \!\!\!\bm{c}_{t}&\!=\!{cross\_{att}}(\bm{a}_{t}, \bm{O}_{t})\!=\!{cross\_{att}}(\bm{a}_{t}, [\bm{o}_{t,k}]^K_{k=1})~\in\mathbb{R}^{d}.\!
    \end{aligned}
     \vspace{-3pt}
\end{equation}
Through cross-attention, \ie, ${cross\_{att}}(\cdot,\cdot)$, action-related visual context $\bm{c}_{t}$ are gathered and compressed into a $d$-dimensional vector.
Then the output of $\mathcal{E}^{r\!}$ is obtained via:
 \vspace{-6pt}
\begin{equation}\small
    \begin{aligned}\label{eq:4}
\!\!\!\!\!\!\!\!\!\!\!\!\text{Ins.~following:}~[\bar{\bm{r}}_{1:t-1}, \bar{\bm{O}}_{t}]&=\!{self_{\!}\_{att}}([\bm{r}_{1:t-1}, \bm{O}_{t}])~\!\in\!\mathbb{R}^{(t-1+K)\times d},\!\!\!\!\!\!\!\!\!\!\!\!\!\!\!\!\!\!\!\!\!\!\! \\
\!\!\!\!\!\!\!\!\!\!\!\!\text{Ins.~generation:}~~~~~~~~~~_{\!}[\bar{\bm{r}}_{1:T}]&=\!{self_{\!}\_{att}}([\bm{r}_{1:T}])~\!\in\!\mathbb{R}^{T\times d}.\!\!
    \end{aligned}
     \vspace{-1pt}
\end{equation}

\noindent\textbf{Language$_{\!}$ Encoder$_{\!}$ $\mathcal{E}^{l}$.}$_{\!}$ $\mathcal{E}^{l\!}$ takes$_{\!}$ input$_{\!}$ either$_{\!}$ the$_{\!}$ complete instruction,$_{\!}$ \ie,$_{\!}$ $X\!=\!\{x_{l}\}^L_{l=1}$, during$_{\!}$ wayfinding;$_{\!}$ or$_{\!}$ previously generated words, \ie, $\{x_1,\cdots_{\!}, x_{l-1}\}$, during instruc- tion$_{\!}$ generation.$_{\!}$ It$_{\!}$ is$_{\!}$ built$_{\!}$ as$_{\!}$ a$_{\!}$ standard$_{\!}$ Transformer$_{\!}$ language$_{\!}$ encoder for contextualized linguistic feature extraction:
     \vspace{-3pt}
\begin{equation}\small
    \begin{aligned}\label{eq:5}
\!\!\!\!\text{Ins.~following:}~~~~~[\bar{\bm{x}}_{1:L}]&\!=\!\mathcal{E}^{l}([x_{1:L}])~\!\in\!\mathbb{R}^{L\times d},\!\!\! \\
\!\!\!\!\text{Ins.~generation:}~[\bar{\bm{x}}_{1:l-1}]&\!=\!\mathcal{E}^{l}([x_{1:l-1}])~\!\in\!\mathbb{R}^{(l-1)\times d},\!\!
    \end{aligned}
         \vspace{-3pt}
\end{equation}
where $\mathcal{E}^{l}$ contains several blocks, each of which has a multi-head self-attention layer and a feed-forward sub-layer~\cite{radford2019language}; position embeddings are omitted for the sake of brevity.

We note that, during the training of the instruction~generation task, \textit{causal future mask}~\cite{vaswani2017attention} is applied to each self-attention layer, ensuring each word token can only attend to the previous ones, and allowing our single model to tackle both instruction following and generation simultaneously.

\noindent\textbf{Route$_{\!}$ Decoder$_{\!}$ $\mathcal{D}^{r\!}$.}$_{\!}$ $\mathcal{D}^{r\!}$ is$_{\!}$ for$_{\!}$ instruction-to-route$_{\!}$ translation.$_{\!}$ Concretely,$_{\!}$ at$_{\!}$ navigation$_{\!}$ step$_{\!}$ $t$,$_{\!}$ $\mathcal{D}^{r\!}$ takes$_{\!}$ input$_{\!}$ the$_{\!}$ complete instruction$_{\!}$ embedding$_{\!}$ $\bar{\bm{x}}_{1:L}$,$_{\!}$ historical$_{\!}$ route$_{\!}$ states$_{\!}$ $\bar{\bm{r}}_{1:t-1}$,$_{\!}$ as well as current observation feature {$\bar{\bm{O}}_t\!=\!\bar{\bm{o}}_{t,1:K}$}, and outputs probability$_{\!}$ distribution$_{\!}$ {$\bm{p}_{t\!}\!\in\!\triangle^{\!K\!\!}$} of$_{\!}$ action$_{\!}$ selection$_{\!}$ {over current $K$ subviews ${{o}}_{t,1:K}$\footnote{Note$_{\!}$ that,$_{\!}$ in$_{\!}$ addition$_{\!}$ to$_{\!}$ the$_{\!}$ $K_{\!}$ action$_{\!}$ subviews,$_{\!}$ \texttt{STOP}$_{\!}$ token$_{\!}$ is$_{\!}$ also$_{\!}$ con- sidered$_{\!}$ here,$_{\!}$ leading$_{\!}$ to$_{\!}$ $K\!+\!1_{\!}$ decision$_{\!}$ choices.$_{\!}$ We$_{\!}$ omit$_{\!}$ \texttt{STOP}$_{\!}$ for$_{\!}$ simplicity.}.}
More specifically, $\mathcal{D}^{r\!}$ is built as a stack of several cross-attention-based blocks for modeling cross-modal relationships. For each block, we have:
         \vspace{-3pt}
\begin{align}\small
	&{\small{[\hat{\bm{r}}_{1:t-1}, \hat{\bm{o}}_{t,1:K}]\!=\!~{cross\_{att}}([\bar{\bm{r}}_{1:t-1}, \bar{\bm{o}}_{t,1:K}], \bar{\bm{x}}_{1:L}),}} \label{eq:6}\\
&{\small{[\bar{\bm{r}}_{1:t-1}, \bar{\bm{o}}_{t,1:K}]\!\leftarrow\!{self_{\!}\_{att}}([\hat{\bm{r}}_{1:t-1}, \hat{\bm{o}}_{t,1:K}]).}} \label{eq:7}
\end{align}
Eq.$_{\!}$~\ref{eq:6}$_{\!}$ obtains$_{\!}$ language-enhanced$_{\!}$ route$_{\!}$ and$_{\!}$ observation$_{\!}$~rep-$_{\!\!}$
 resentations, \ie,  $\hat{\bm{r}}_{1:t-1}$ and $\hat{\bm{o}}_{t,1:K}$, through cross-attention. Eq.$_{\!}$~\ref{eq:7}$_{\!}$ adopts$_{\!}$ self-attention$_{\!}$ to$_{\!}$ model$_{\!}$ temporal$_{\!}$ dependencies among historical route states $\hat{\bm{r}}_{1:t-1}$, and capture  the correlations between $\hat{\bm{r}}_{1:t-1}$ and current observation $\hat{O}_{t}\!=\!\hat{\bm{o}}_{t,1:K}$.

After several $\mathcal{D}^{r\!}$ decoder blocks, the action probability over the $K$ subviews ${o}_{t,1:K}$ is given as:
   \vspace*{-2pt}
\begin{equation}\small
    \begin{aligned}\label{eq:8}
\bm{p}_{t} = softmax(\{\mathcal{F}^r(\bar{\bm{o}}_{k})\}^K_{k=1})~~\in\triangle^{\!K},
    \end{aligned}
       \vspace*{-3pt}
\end{equation}
where$_{\!}$ $\mathcal{F}^r\!\!:{\!}\mathbb{R}^d\!\rightarrow\!\mathbb{R}_{\!}$ is$_{\!}$ a$_{\!}$ two-layer$_{\!}$ feed-forward$_{\!}$ network$_{\!}$ for action score mapping, as in~\cite{chen2021history,zhao2022target}.

\noindent\textbf{Language$_{\!}$ Decoder$_{\!}$ $\mathcal{D}^{l\!}$.} $\mathcal{D}^{l\!}$ is$_{\!}$ for$_{\!}$ route-to-instruction$_{\!}$ trans- lation.$_{\!}$ Concretely,$_{\!}$ at$_{\!}$ instruction$_{\!}$ generation$_{\!}$ step$_{\!}$ $l$,$_{\!}$ $\mathcal{D}^{l\!}$ takes$_{\!}$ input$_{\!}$ the$_{\!}$ full$_{\!}$ route$_{\!}$ states$_{\!}$ $\bar{\bm{r}}_{1:T}$, and the embeddings of pre- viously$_{\!}$ generated$_{\!}$ instruction$_{\!}$ words$_{\!}$ $\bar{\bm{x}}_{1:{l-1}}$,$_{\!}$ and$_{\!}$ outputs$_{\!}$ pro- bability$_{\!}$ distribution$_{\!}$ $\bm{q}_{l\!}\!\in\!\!\triangle^{\!M\!\!}$ of$_{\!}$ word$_{\!}$ selection$_{\!}$ over$_{\!}$ a$_{\!}$ pre-defined$_{\!}$ vocabulary$_{\!}$ with$_{\!}$ $M_{\!}$ words.$_{\!}$ Analogous$_{\!}$ to$_{\!}$ $\mathcal{D}^{r}$,$_{\!}$ $\mathcal{D}^{l\!}$ has several$_{\!}$ cross-attention$_{\!}$ based$_{\!}$ blocks.$_{\!}$ Each$_{\!}$ block$_{\!}$ is$_{\!}$ given$_{\!}$ as:
   \vspace*{-3pt}
\begin{align}\small
	&{\small{\hat{\bm{x}}_{1:l-1}\!=\!~{cross\_{att}}(\bar{\bm{x}}_{1:l-1}, \bar{\bm{r}}_{1:T}),}} \label{eq:9}\\
&{\small{\bar{\bm{x}}_{1:l-1}\!\leftarrow\!{causal\_self_{\!}\_{att}}(\hat{\bm{x}}_{1:l-1}).}} \label{eq:10}
\end{align}
Eq.~\ref{eq:9} lets the text attend to the route context. In Eq.~\ref{eq:10}, we adopt the \textit{causally-masked self-attention}, instead of normal, bi-directional self-attention, to force $\mathcal{D}^{l\!}$  to ``attend-ahead'', which is needed for autoregressive inference.

After several $\mathcal{D}^{l\!}$ decoder blocks, the probability over the $M$-word vocabulary is given as:
   \vspace*{-2pt}
\begin{equation}\small
    \begin{aligned}\label{eq:11}
\bm{q}_{l} = softmax(\mathcal{F}^l( \bar{\bm{x}}_{l-1} ))~~\in\triangle^{\!M},
    \end{aligned}
       \vspace*{-3pt}
\end{equation}
where$_{\!}$ $\mathcal{F}^r\!\!:{\!}\mathbb{R}^d\!\rightarrow\!\mathbb{R}^{M\!}$ is$_{\!}$ a$_{\!}$ two-layer$_{\!}$ feed-forward$_{\!}$ network$_{\!}$ for the prediction of the word score distribution.

\subsection{Network Training}\label{sec:nt}
All the modules of \textsc{Lana}, \ie, two unimodal encoders $\mathcal{E}^{r\!}$ and $\mathcal{E}^{l}$, as well as two  multimodal decoders $\mathcal{D}^{r\!}$ and $\mathcal{D}^{l\!}$, are jointly end-to-end learned, by optimizing the training~objec- tives of instruction following and generation.

\noindent\textbf{Instruction$_{\!}$ Generation.$_{\!}$} For$_{\!}$ each$_{\!}$ instruction-route$_{\!}$ training pair$_{\!}$ $(X, R)$,$_{\!}$ where$_{\!}$ $X\!=\!x_{1:L\!}$ and$_{\!}$ $R\!=\!r_{1:T}$, \textsc{Lana} learns instruction generation by predicting $x_l$ based on the full route $R_{\!}$ and$_{\!}$ proceeding$_{\!}$ reference$_{\!}$ words$_{\!}$ $x_{0:l-1\!}$.$_{\!}$ We$_{\!}$ append$_{\!}$ two special$_{\!}$ tokens$_{\!}$ to$_{\!}$ $X$,$_{\!}$ \ie,$_{\!}$ $x_{0}\!=\![\texttt{BOS}]_{\!}$ and$_{\!}$ $x_{L+1}\!=\![\texttt{EOS}]$,~res- pectively$_{\!}$ indicating$_{\!}$ the$_{\!}$ start$_{\!}$ and$_{\!}$ end$_{\!}$ of$_{\!}$ the$_{\!}$ instruction sen- tence.$_{\!}$ To$_{\!}$ generate$_{\!}$ word$_{\!}$ $x_l$,$_{\!}$ \textsc{Lana}$_{\!}$ respectively$_{\!}$ feeds$_{\!}$ $R_{\!}$ and$_{\!}$ $x_{0:l-1\!}$ into$_{\!}$ $\mathcal{E}^{r\!}$ and$_{\!}$ $\mathcal{E}^{l\!}$ for$_{\!}$ unimodal$_{\!}$ encoding$_{\!}$ (\textit{cf}.$_{\!}$~Eq.$_{\!}$~\ref{eq:4}\&\ref{eq:5}). Conditioned$_{\!}$ on$_{\!}$ the$_{\!}$ route$_{\!}$ and linguistic$_{\!}$~embeddings,$_{\!}$ \ie,$_{\!}$ $\bar{\bm{r}}_{1:T}$ and$_{\!}$ $\bar{\bm{x}}_{1:l-1}$,$_{\!}$ $\mathcal{D}^{l\!}$ gives the word probability $\bm{q}_l$ (\textit{cf}.$_{\!}$~Eq.$_{\!}$~\ref{eq:11}).
The training objective of instruction generation, formulated as the language modeling loss, can be written as:
\vspace*{-2pt}
\begin{equation}\small
    \begin{aligned}\label{eq:12}
    \mathcal{L}^{g} = -\sum_{l=1}^{L+1}{\log}(p(x_l|x_{0:l-1}, R)) = -\sum_{l=1}^{L+1}{\log}(\bm{q}_l(x_l)),
    \end{aligned}
    \vspace*{-3pt}
\end{equation}
where  $\bm{q}_l(x_l)\!\in\![0,1]$ is the probability of word $x_l$. \textsc{Lana}~is

\noindent trained to minimize the negative
log-likelihood of the reference instruction$_{\!}$ words.$_{\!}$ Teacher-forcing$_{\!}$~\cite{williams1989learning}$_{\!}$ is$_{\!}$ used$_{\!}$ here$_{\!}$ to$_{\!}$ enable$_{\!}$ the parallel$_{\!}$ text$_{\!}$ input.$_{\!}$ Worth$_{\!}$ mentioning$_{\!}$ is$_{\!}$ that,$_{\!}$ existing VLN pretraining methods \cite{guhur2021airbert,hao2020towards,guhur2021airbert,chen2021history,qiao2022hop} rely on the masked language modeling (MLM) strategy. Since MLM only predicts a small portion (typically 15\%) of input words during each training iteration, it is less efficient for large-scale pretraining data, as pointed out by many recent literature in general vision-language pretraining~\cite{hu2022scaling,brown2020language,chowdhery2022palm}.

\noindent\textbf{Instruction$_{\!}$ Following.$_{\!}$} For$_{\!}$ each$_{\!}$ training pair$_{\!}$ $(X, R)$,$_{\!}$ where $X\!=\!x_{1:L\!}$ and$_{\!}$ $R\!=\!r_{1:T}\!=\!(O_{t}, a_{t})_{1:T}$,$_{\!}$ \textsc{Lana}$_{\!}$ concurrently
learns instruction following by predicting $a_t$ based on the full$_{\!}$ instruction$_{\!}$ $X$,$_{\!}$ history$_{\!}$ from$_{\!}$ expert$_{\!}$ demonstration$_{\!}$ $r_{1:t-1}$,
and the current percept $O_t$. Specifically, \textsc{Lana}$_{\!}$ respectively
feeds$_{\!}$ $X_{\!}$ and$_{\!}$ $\{r_{1:t-1\!}, O_t\}$ into$_{\!}$ $\mathcal{E}^{l\!}$ and$_{\!}$ $\mathcal{E}^{r}$ (\textit{cf}.$_{\!}$~Eq.$_{\!}$~\ref{eq:4}\&\ref{eq:5}). Conditioned on the output unimodal encodings, \ie, $\bar{\bm{x}}_{1:L}$ and$_{\!}$ $[\bar{\bm{r}}_{1:t-1}, \bar{\bm{O}}_{t}]$, $\mathcal{D}^{r\!}$ gives the action probability $\bm{p}_t$ (\textit{cf}.$_{\!}$~Eq.$_{\!}$~\ref{eq:8}). The training objective of instruction following is to mini- mize$_{\!}$ the$_{\!}$ negative$_{\!}$ log-likelihood$_{\!}$ of$_{\!}$ the$_{\!}$ target$_{\!}$ view$_{\!}$ action$_{\!}$ $a_t$:
\vspace*{-3pt}
\begin{equation}\small
    \begin{aligned}\label{eq:13}
    \!\!\mathcal{L}^{f}\!=\!-\!\sum_{t=1}^{T}\log(p(a_t|r_{0:t-1}, O_t, X))\!=\!-\!\sum_{t=1}^{T}\log(\bm{p}_t(a_t)).\!\!
    \end{aligned}
    \vspace*{-3pt}
\end{equation}

$_{\!}$\textsc{Lana}$_{\!}$ is$_{\!}$ end-to-end$_{\!}$ learned$_{\!}$ with$_{\!}$ the$_{\!}$ two$_{\!}$ training$_{\!}$ targets (\textit{cf}.$_{\!}$~Eq.${\!}$~\ref{eq:12}\&\ref{eq:13})$_{\!}$ during$_{\!}$ both$_{\!}$ pretraing$_{\!}$ and$_{\!}$ fine-tuning phases. Note$_{\!}$ that$_{\!}$ the$_{\!}$ encoders$_{\!}$ $\mathcal{E}^{r\!}$ and$_{\!}$ $\mathcal{E}^{l\!}$ receive$_{\!}$ the$_{\!}$ supervision$_{\!}$ sig- nals$_{\!}$ from$_{\!}$ both$_{\!}$ instruction$_{\!}$ generation$_{\!}$ (\textit{cf}.$_{\!}$~Eq.$_{\!}$~\ref{eq:12})$_{\!}$ and following (\textit{cf}.$_{\!}$~Eq.$_{\!}$~\ref{eq:13}). Moreover, such a joint learning framework grants \textsc{Lana} improved interpretability -- \textsc{Lana} can be viewed as a navigator born with a language explainer  $\mathcal{D}^{l}$.

\subsection{Implementation Details}

\sssection{Network$_{\!}$ Architecture.$_{\!}$} The$_{\!}$ route$_{\!}$ $\mathcal{E}^{r\!}$ and$_{\!}$ language$_{\!}$ $\mathcal{E}^{l\!}$ enco- ders$_{\!}$ respectively$_{\!}$ have$_{\!}$ one and nine layers, and the decoders $\mathcal{D}^{r\!\!}$ and$_{\!}$ $\mathcal{D}^{l\!}$ both$_{\!}$ have$_{\!}$ four$_{\!}$ blocks.$_{\!}$ The$_{\!}$ feature$_{\!}$ dimension$_{\!}$ is$_{\!}$ set$_{\!}$ as $d\!=\!768$. The orientation feature $\bm{\theta}_k$ of view $o_k$ (\textit{cf}.$_{\!}$~Eq.$_{\!}$~\ref{eq:1}) is defined as: $\bm{\theta}_{k}\!=\!(\cos\varphi_{k}, \sin\varphi_{k}, \cos\phi_{k}, \sin\phi_{k})$, where $\varphi$ and  $\phi$ are the angles of heading and elevation, respectively.

\noindent\textbf{Training.} Following recent VLN practice \cite{majumdar2020improving,hao2020towards,guhur2021airbert,chen2021history,qiao2022hop}, the pretraining and fine-tuning paradigm is adopted:
\begin{itemize}[leftmargin=*]
	\setlength{\itemsep}{0pt}
	\setlength{\parsep}{-2pt}
	\setlength{\parskip}{-0pt}
	\setlength{\leftmargin}{-10pt}
	\vspace{-5pt}
	\item \textit{Pretraining}:$_{\!}$ With$_{\!}$ the$_{\!}$ two$_{\!}$ training$_{\!}$ objectives$_{\!}$ (\textit{cf}.$_{\!}$~Eq.$_{\!}$~\ref{eq:12}\& \ref{eq:13}),$_{\!}$  \textsc{Lana}$_{\!}$ is$_{\!}$ pretrained$_{\!}$  on$_{\!}$ offline-sampled$_{\!}$ instruction-route$_{\!}$ pairs$_{\!}$ from$_{\!}$ PREVALENT$_{\!}$~\cite{hao2020towards},$_{\!}$ including$_{\!}$ 104K$_{\!}$ ori- ginal R2R samples and 6482K synthesized ones. \textsc{Lana} is trained for 100K iterations, using Adam optimizer~\cite{kingma2015adam} with 1e-4 learning rate, and$_{\!}$ $N\!=\!128_{\!}$ batch$_{\!}$ size.$_{\!}$ 
    \item \textit{Fine-tuning}: Then we fine-tune \textsc{Lana} on different VLN datasets,$_{\!}$ still$_{\!}$ using$_{\!}$ our$_{\!}$ two$_{\!}$ training$_{\!}$ tasks$_{\!}$ (\textit{cf}.$_{\!}$~Eq.$_{\!}$~\ref{eq:12}\&\ref{eq:13}). Following the standard protocol~\cite{hong2020recurrent,wang2022counterfactual,hong2020language,wang2019reinforced}, the training of instruction following is based on the mixture of \textit{imitation learning} and \textit{reinforcement learning}. In this stage, we set the learning rate to 1e-5 and batch size to 8. 
	\vspace{-8pt}
\end{itemize}
We use four NVIDIA Tesla A100 GPUs for network training, and sample only one training task for each mini-batch. 

\noindent\textbf{Inference.$_{\!}$} Once$_{\!}$ trained,$_{\!}$ \textsc{Lana} is$_{\!}$ capable$_{\!}$ of$_{\!}$ both$_{\!}$ following and verbalizing navigation instructions with only one single model instance, without any architectural change. Specifically, for instruction following, greedy search, \ie, selecting the action with the highest probability at each prediction step, is adopted and terminated when \texttt{STOP} is chosen. For instruction generation, the sentence is predicted in an autoregressive manner, \ie,  generating one word at a time until \texttt{EOS} is chosen, conditioned on previous generated ones.

\begin{figure*}
\begin{minipage}{\textwidth}
  \begin{minipage}[t]{0.495\textwidth}
    \begin{threeparttable}
        \resizebox{1\textwidth}{!}{
		\setlength\tabcolsep{2pt}
            \renewcommand\arraystretch{1.03}
    \begin{tabular}{|rl||cccc|cccc|}
\hline \thickhline
   ~ & &\multicolumn{4}{c|}{R2R \texttt{val} \texttt{unseen}} & \multicolumn{4}{c|}{R2R \texttt{test} \texttt{unseen}} \\
  \cline{3-10}\cline{3-10}\cline{3-10}
\multicolumn{2}{|c||}{\multirow{-2}{*}{Methods}} &\textbf{\texttt{SR}}\!~$\uparrow$ &\textbf{\texttt{SPL}}\!~$\uparrow$ &\texttt{OR}\!~$\uparrow$    &\texttt{TL}\!~$\downarrow$  &\textbf{\texttt{SR}}\!~$\uparrow$  &\textbf{\texttt{SPL}}\!~$\uparrow$ &\texttt{OR}\!~$\uparrow$  &\texttt{TL}\!~$\downarrow$\\
\hline
\hline
BT-follower~\cite{fried2018speaker}\!&\pub{NeurIPS2018} & 36   & - & 45 & - & 35   & 28 & 44  & 14.8\\
EDrop-follower~\cite{tan2019learning}\!&\pub{NAACL2019} & 52  &48 & - & {10.7} & 51   & 47 & 59 & {11.7} \\
AuxRN~\cite{zhu2019vision}\!&\pub{CVPR2020} &55   & {50} & 62 & - & 55  & 51 &62  & - \\
PREVALENT~\cite{hao2020towards}\!&\pub{CVPR2020} &58  &53 &- &10.2 &54   &51  &- &10.5 \\
VLN$\circlearrowright$BERT~\cite{hong2020recurrent}\!&\pub{CVPR2021} &63 &57 &- &12.0 &63 &57 &- &12.4 \\
AirBERT~\cite{guhur2021airbert}\!&\pub{ICCV2021} &62  &56 &- &11.8 &62   &57 &- &12.4 \\
HAMT~\cite{chen2021history}\!&\pub{NeurIPS2021} &65 &59 &-  &11.9 &63  &58 &- &12.7 \\
 HOP~\cite{qiao2022hop}\!&\pub{CVPR2022}  &64  &57 &- &12.3 &64 &59 &- &12.7  \\
  \hline
  \textsc{Lana}$_{st}~$&\footnotesize{(\textbf{\texttt{ours}})} & 66  & 60 & 73 & 11.9 & 64  & 59 & 71 & 12.4 \\
  \textsc{Lana}$_{mt}$&\footnotesize{(\textbf{\texttt{ours}})}  & \textbf{68}  & \textbf{62} & \textbf{76} & 12.0 & \textbf{65}  & \textbf{60} & \textbf{71} & 12.6 \\
\hline
\end{tabular}
    }
\end{threeparttable}
    \vspace*{-8pt}
    \makeatletter\def\@captype{table}\makeatother\captionsetup{font=small}\caption{\small{Quantitative comparison results (\S\ref{sec:ex_if}) for \textbf{instruction following} on R2R$_{\!}$~\cite{anderson2018vision}. `$-$': unavailable statistics.\label{table:IFr2r}}  }
  \end{minipage}
  \begin{minipage}[t]{0.005\textwidth}
  ~~~~~~
  \end{minipage}
  \begin{minipage}[t]{0.495\textwidth}
    \begin{threeparttable}
        \resizebox{1\textwidth}{!}{
		\setlength\tabcolsep{5.3pt}
            \renewcommand\arraystretch{1.03}
    \begin{tabular}{|rl||ccccc|}
    \hline \thickhline
    ~ &&\multicolumn{5}{c|}{R4R \texttt{val} \texttt{unseen}} \\
    \cline{3-7}\cline{3-7}\cline{3-7}
    \multicolumn{2}{|c||}{\multirow{-2}{*}{Methods}} &\textbf{\texttt{CLS}}~$\uparrow$ &\texttt{nDTW}~$\uparrow$ &\texttt{SDTW}~$\uparrow$ &\texttt{SR}~$\uparrow$ &\texttt{TL}~$\downarrow$  \\
    \hline
    \hline
    BT-follower~\cite{fried2018speaker}\!\!&\!\!\!\pub{NeurIPS2018} & 30 & - & - & 24  & 19.9   \\
    RCM~\cite{wang2019reinforced}\!\!&\!\!\!\pub{CVPR2019}  & 35 & 30 & 13 & 26  & 28.5  \\
    PTA~\cite{landi2020perceive}\!\!&\!\!\!\pub{NeurIPS2021} & 37 & 32 & 10 & 24 & 17.7   \\
    EDrop-follower~\cite{tan2019learning}\!\!&\!\!\!\pub{NAACL2019}  & 34 & - & 9 & 29 & 27.0    \\
    OAAM~\cite{qi2020object}\!\!&\!\!\!\pub{ECCV2020}  & 40 & -  & 11 & 31 & 13.8  \\
     ActiveVLN~\cite{wang2020active}\!\!&\!\!\!\pub{ECCV2020}   &59 &44 &22 &32 &19.7 \\
     EGP~\cite{deng2020evolving}\!\!&\!\!\!\pub{NeurIPS2020} & 44 & 37 & 18 & 30 & 18.3   \\
    HAMT~\cite{chen2021history}\!\!&\!\!\!\pub{NeurIPS2021} &57.7 &50.3 &31.8 &44.6  &-  \\
  \hline
  \textsc{Lana}$_{st}~$\!\!&\!\!\footnotesize{(\textbf{\texttt{ours}})} & 58.6 & 51.9 & 31.4 & 43.0 & 22.7    \\
  \textsc{Lana}$_{mt}$\!\!&\!\!\footnotesize{(\textbf{\texttt{ours}})}  & \textbf{59.7} & \textbf{52.3} & 31.7 & 43.2  & 22.1  \\
\hline
    \end{tabular}
    }
\end{threeparttable}
    \vspace*{-8pt}
    \makeatletter\def\@captype{table}\makeatother\captionsetup{font=small}\caption{\small{$_{\!\!}$Quantitative$_{\!}$ comparison$_{\!}$ results$_{\!}$ (\S\ref{sec:ex_if})$_{\!}$ for$_{\!}$ \textbf{instruction$_{\!}$ fol- lowing} on R4R~\cite{jain2019stay}.\label{table:IFr4r}}  }
  \end{minipage}
    \end{minipage}
  \vspace*{-8pt}
\end{figure*}

\begin{table*}[t]
\centering
\resizebox{0.9\textwidth}{!}{
\setlength\tabcolsep{7pt}
\renewcommand\arraystretch{1.03}
\begin{tabular}{|rl||cccccc|cccccc|}
\hline \thickhline
   ~ & & \multicolumn{6}{c|}{REVERIE \texttt{val} \texttt{unseen}} & \multicolumn{6}{c|}{REVERIE \texttt{test} \texttt{unseen}} \\
  \cline{3-14}\cline{3-14}\cline{3-14}\cline{3-14}
\multicolumn{2}{|c||}{\multirow{-2}{*}{Methods}}
&\textbf{\texttt{SR}}$_{\!}$~$\uparrow$  &\textbf{\texttt{SPL}}$_{\!}$~$\uparrow$ &\texttt{OR}~$\uparrow$  &\texttt{TL}~$\downarrow$ &\texttt{RGS}$_{\!}$~$\uparrow$&\texttt{RGSPL}$_{\!}$~$\uparrow$
&\textbf{\texttt{SR}}$_{\!}$~$\uparrow$ &\textbf{\texttt{SPL}}$_{\!}$~$\uparrow$ &\texttt{OR}$_{\!}$~$\uparrow$  &\texttt{TL}~$\downarrow$ &\texttt{RGS}$_{\!}$~$\uparrow$&\texttt{RGSPL}$_{\!}$~$\uparrow$\\
\hline
\hline
RCM~\cite{wang2019reinforced}\!\!&\!\!\!\!\pub{CVPR2019}  &9.29&6.97&14.23&11.98&4.89&3.89&7.84&6.67&11.68&10.60&3.67&3.14\\
VLN$\circlearrowright$BERT~\cite{hong2020recurrent}\!\!&\!\!\!\!\pub{CVPR2021}&30.67&24.90&35.02&16.78&18.77&15.27&29.61&23.99&32.91&15.86&16.50&13.51\\
AirBERT~\cite{guhur2021airbert}\!\!&\!\!\!\!\pub{ICCV2021}&27.89&21.88&34.51&18.71&18.23&14.18&30.28&23.61&34.20&17.91&16.83&13.28\\
HAMT~\cite{chen2021history}\!\!&\!\!\!\!\pub{NeurIPS2021} &32.95 &30.20  &36.84  &14.08 &18.92 &17.28  &30.40 &26.67  &33.41 &13.62 &14.88 &13.08\\
HOP~\cite{qiao2022hop}\!\!&\!\!\!\!\pub{CVPR2022} & 30.39 & 25.10 & 35.30  & 17.16 & 18.23 &15.31 & 29.12 &23.37 & 32.26  & 17.05 &17.13 &13.90 \\
  \hline
  \textsc{Lana}\!\!&\!\!\!\!\footnotesize{(\textbf{\texttt{ours}})} & \textbf{34.00} & 29.26 &  \textbf{38.54}  & 16.28 & \textbf{19.03} & 16.18& \textbf{33.50} & \textbf{26.89} & \textbf{36.41}  & 16.75 & \textbf{17.53} & \textbf{14.25}  \\
\hline
\end{tabular}
}
\captionsetup{font=small}
\caption{\small{Quantitative comparison results (\S\ref{sec:ex_if}) for \textbf{instruction following} on REVERIE$_{\!}$~\cite{2020REVERIE}.}}    \label{table:IFrev}
\vspace*{-8pt}
\end{table*}

\begin{table*}[!bth]
	\centering
			\resizebox{1\textwidth}{!}{
			\setlength\tabcolsep{3pt}
			\renewcommand\arraystretch{1.0}
	\begin{tabular}{|rl||cccccc|cccccc|}
	\hline \thickhline
	~ & & \multicolumn{6}{c|}{R2R \texttt{val} \texttt{seen}} & \multicolumn{6}{c|}{R2R \texttt{val} \texttt{unseen}}\\
	\cline{3-14}\cline{3-14}
	\multicolumn{2}{|c||}{\multirow{-2}{*}{Methods}}
	&\textbf{\texttt{SPICE}}\!~$\uparrow$ &\texttt{Bleu-1}\!~$\uparrow$ &\texttt{Bleu-4}\!~$\uparrow$ &\texttt{CIDEr}\!~$\uparrow$ &\texttt{Meteor}\!~$\uparrow$ &\texttt{Rouge}\!~$\uparrow$  &\textbf{\texttt{SPICE}}\!~$\uparrow$ &\texttt{Bleu-1}\!~$\uparrow$ &\texttt{Bleu-4}\!~$\uparrow$ &\texttt{CIDEr}\!~$\uparrow$ &\texttt{Meteor}\!~$\uparrow$ &\texttt{Rouge}\!~$\uparrow$  \\
	\hline
	\hline
	BT-speaker$_{\!}$~\cite{fried2018speaker}&\!\!\pub{NeurIPS2018}    &0.203 &0.537 &0.155 &0.121 &0.233 &0.350   &0.188 &0.522 &0.142 &0.114 &0.228 &0.346 \\
    EDrop-speaker$_{\!}$~\cite{tan2019learning}&\!\!\pub{NAACL2019}   &0.202 &- & 0.245 &0.493 &0.228 &0.467   &0.181 & - &0.237 &0.422 &0.225 &0.458 \\
	VLS~\cite{agarwal2019visual}&\!\!\pub{CVPRW2019}   &0.214 &0.549 &0.157 &0.137 &0.228 &0.352   &0.197 &0.548 &0.159 &0.132 &0.231 &0.357  \\
	CCC-speaker$_{\!}$~\cite{wang2022counterfactual}&\!\!\pub{CVPR2022}  &{0.231}  &{0.728} & 0.287 & 0.543 & {0.236} & {0.493}  &{0.214} &{0.708} & 0.272 & 0.461 &{0.231} &{0.477} \\
  \hline
  \textsc{Lana}$_{st}~$&\!\!\footnotesize{(\textbf{\texttt{ours}})} & 0.251  & 0.743 & 0.305 & 0.522 & 0.243 & 0.502 & 0.223 & 0.722 & 0.287 & 0.433 & 0.235 & 0.490  \\
  \textsc{Lana}$_{mt}$&\!\!\footnotesize{(\textbf{\texttt{ours}})}  & \textbf{0.256} &  \textbf{0.759} &  \textbf{0.314} & 0.533 &  \textbf{0.245} &  \textbf{0.503} &  \textbf{0.226} &  \textbf{0.736} &  \textbf{0.298} & 0.457 &  \textbf{0.238} &  \textbf{0.498}  \\
\hline
	\end{tabular}
	}
		\vspace*{-2pt}
	\captionsetup{font=small}
		\caption{\small{Quantitative comparison results (\S\ref{sec:ex_ig}) for \textbf{instruction generation} on R2R~\cite{anderson2018vision}.  }}
		\label{table:R2Rig}
	\vspace*{-12pt}
\end{table*}

\section{Experiment}\label{sec:ex}
\vspace*{-2pt}
We$_{\!}$ evaluate$_{\!}$ \textsc{Lana}$_{\!}$ for$_{\!}$ both$_{\!}$ instruction$_{\!}$ following$_{\!}$  (\S\ref{sec:ex_if}) and$_{\!}$~generation (\S\ref{sec:ex_ig}) tasks, followed by a series of diagnostic experiments (\S\ref{sec:abl}) and qualitative studies (\S\ref{sec:ve}).

For each task, we give scores of two versions of \textsc{Lana}:
\begin{itemize}[leftmargin=*]
	\setlength{\itemsep}{0pt}
	\setlength{\parsep}{-2pt}
	\setlength{\parskip}{-0pt}
	\setlength{\leftmargin}{-10pt}
	\vspace{-5pt}
	\item \textsc{Lana}$_{mt}$:$_{\!}$ jointly$_{\!}$ learn$_{\!}$ the$_{\!}$ two$_{\!}$ target$_{\!}$ tasks$_{\!}$ throughout$_{\!}$~pre- training$_{\!}$ and$_{\!}$ fine-tuning.$_{\!}$ Thus,$_{\!}$ such$_{\!}$ multi-task$_{\!}$ version$_{\!}$ only$_{\!}$ has$_{\!}$~one single$_{\!}$ agent$_{\!}$ instance, tested on the two tasks.
	\item \textsc{Lana}$_{st}$:$_{\!}$ jointly$_{\!}$ pre-train$_{\!}$ on$_{\!}$ the$_{\!}$ two$_{\!}$ tasks,$_{\!}$ but$_{\!}$ fine-tune$_{\!}$ on each$_{\!}$ task$_{\!}$ individually.$_{\!}$ There$_{\!}$ are$_{\!}$ two$_{\!}$ single-task$_{\!}$ agent$_{\!}$ instances; each is only tested on the corresponding task.
	\vspace{-3pt}
\end{itemize}

\begin{table*}[t]
	\centering
			\resizebox{1\textwidth}{!}{
			\setlength\tabcolsep{2pt}
			\renewcommand\arraystretch{1.0}
	\begin{tabular}{|rl||cccccc|cccccc|}
	\hline \thickhline
	~ & & \multicolumn{6}{c|}{R4R \texttt{val} \texttt{seen}} & \multicolumn{6}{c|}{R4R \texttt{val} \texttt{unseen}}\\
	\cline{3-14}\cline{3-14}\cline{3-14}
	\multicolumn{2}{|c||}{\multirow{-2}{*}{Methods}}
	&\textbf{\texttt{SPICE}}\!~$\uparrow$ &\texttt{Bleu-1}\!~$\uparrow$ &\texttt{Bleu-4}\!~$\uparrow$ &\texttt{CIDEr}\!~$\uparrow$ &\texttt{Meteor}\!~$\uparrow$ &\texttt{Rouge}\!~$\uparrow$ &\textbf{\texttt{SPICE}}\!~$\uparrow$  &\texttt{Bleu-1}\!~$\uparrow$ &\texttt{Bleu-4}\!~$\uparrow$ &\texttt{CIDEr}\!~$\uparrow$ &\texttt{Meteor}\!~$\uparrow$ &\texttt{Rouge}\!~$\uparrow$  \\
	\hline
	\hline
	BT-speaker$_{\!}$~\cite{fried2018speaker}&\!\!\pub{NeurIPS2018}   & 0.164 & 0.691 & 0.223 & 0.099 & 0.213 &0.453 & 0.207 & 0.387 & 0.088 & 0.139& 0.172 & 0.359	\\
    EDrop-speaker$_{\!}$~\cite{tan2019learning}&\!\!\pub{NAACL2019}   & 0.209 & 0.750 & 0.281 & 0.216 & 0.245 & 0.473 & 0.218 & 0.433 & 0.106  & 0.200	& 0.187 & 0.363	 \\
	CCC-speaker$_{\!}$~\cite{wang2022counterfactual}&\!\!\pub{CVPR2022} & 0.219 & 0.758 & 0.312 & 0.245 & 0.252 & 0.480 & 0.233 & 0.403 & 0.115 & 0.206 & 0.193 & 0.365   \\
  \hline
  \textsc{Lana}$_{st}~$&\!\!\footnotesize{(\textbf{\texttt{ours}})} & 0.237 & 0.768 & 0.327 & 0.264 & \textbf{0.265} & 0.483 & 0.259 & 0.437 & 0.123 & 0.216 & 0.199 & 0.375 \\
  \textsc{Lana}$_{mt}$&\!\!\footnotesize{(\textbf{\texttt{ours}})} & \textbf{0.245} & \textbf{0.772} & \textbf{0.333} & \textbf{0.287} & 0.261 & \textbf{0.484} & \textbf{0.262} & \textbf{0.443} & \textbf{0.128} & \textbf{0.231} & \textbf{0.200} & \textbf{0.376} \\
\hline
	\end{tabular}
	}
		\vspace*{-2pt}
	\captionsetup{font=small}
		\caption{\small{Quantitative comparison results (\S\ref{sec:ex_ig}) for \textbf{instruction generation} on R4R~\cite{jain2019stay}.  }}
		\label{table:R4Rig}
	\vspace*{-6pt}
\end{table*}

\begin{table*}[!bth]
	\centering
	\resizebox{1\textwidth}{!}{
	\setlength\tabcolsep{4pt}
	\renewcommand\arraystretch{1.0}
	\begin{tabular}{|c|cc|cc||cccc|cccccc|}
	\hline \thickhline
	\multirow{3}{*}{\#}&\multicolumn{2}{c|}{Pretraining}&\multicolumn{2}{c||}{Fine-tuning} & \multicolumn{4}{c|}{Instruction Following} & \multicolumn{6}{c|}{Instruction Generaion}\\
\cline{2-15}\cline{2-15}
&\tabincell{c|}{\textit{Instruction}\\\textit{Following}} &\tabincell{c}{\textit{Instruction}\\\textit{Generation}} &\tabincell{c|}{\textit{Instruction}\\\textit{Following}} &\tabincell{c}{\textit{Instruction}\\\textit{Generation}}	&\textbf{\texttt{SR}}~$\uparrow$ &\textbf{\texttt{SPL}}~$\uparrow$ &\texttt{OR}~$\uparrow$  &\texttt{TL}~$\downarrow$
&\textbf{\texttt{SPICE}}~$\uparrow$  &\texttt{Bleu-1}~$\uparrow$ &\texttt{Bleu-4}~$\uparrow$ &\texttt{CIDEr}~$\uparrow$ &\texttt{Meteor}~$\uparrow$ &\texttt{Rouge}~$\uparrow$ \\
	\hline	\hline
1&&&\cmark& & 52.1 & 48.3 & 59.3 & 11.2 &- &- &- &- &- &- \\
2&&&&\cmark & - & - & -& - & 0.178 & 0.692 & 0.241 & 0.321 & 0.216 & 0.463 \\
3&&&\cmark&\cmark & 53.1 & 48.9 & 60.9 & 11.6 & 0.182 & 0.704 & 0.245 & 0.304 & 0.219 & 0.469 \\
\hline
4&\cmark&&\cmark& & 61.3 & 55.4 & 69.7 & 12.0 &- &- &- &- &- &- \\
5&&\cmark&&\cmark & - & - & - & - & 0.215 & 0.718 & 0.255 & 0.378 & 0.230 & 0.472 \\
\hline
6&\cmark&\cmark&\cmark& & 65.7  & 59.9 & 73.4 & 11.9 &- &- &- &- &- &- \\
7&\cmark&\cmark&&\cmark &- &- &- &- & 0.223 & 0.722 & 0.287 & 0.433 & 0.235 & 0.490 \\
8&\cmark&\cmark&\cmark&\cmark & \textbf{67.9}  & \textbf{61.6} & \textbf{75.7} & 12.0 &\textbf{0.226} &  \textbf{0.736} &  \textbf{0.298} & \textbf{0.457} &  \textbf{0.238} &  \textbf{0.498} \\
\hline
	\end{tabular}
	}
		\vspace*{-2pt}
	\captionsetup{font=small}
		\caption{\small{\textbf{Ablation study} (\S\ref{sec:abl}) on R2R \texttt{val} \texttt{unseen}}~\cite{anderson2018vision}.  }
		\label{table:R2Rab}
	\vspace*{-12pt}
\end{table*}

\noindent We$_{\!}$ collect$_{\!}$ here$_{\!}$ key$_{\!}$ observations$_{\!}$ from$_{\!}$ our$_{\!}$ subsequent$_{\!}$ expe- riments:$_{\!}$ \textbf{i)}$_{\!}$ \textsc{Lana}$_{\!}$ performs$_{\!}$ comparable,$_{\!}$ or$_{\!}$ even$_{\!}$ better$_{\!}$ than prior$_{\!}$ tasks-specific$_{\!}$ agents;$_{\!}$ \textbf{ii)}$_{\!}$ \textsc{Lana}$_{mt\!}$ outperforms$_{\!}$ \textsc{Lana}$_{st\!}$ with more efficient$_{\!}$ parameter$_{\!}$ utilization;$_{\!}$ \textbf{iii)}$_{\!}$ \textsc{Lana}$_{\!}$ can$_{\!}$ pro- vide test-time behavioral interpretation by verbalizing des- criptions of its navigation routes; and \textbf{iv)}$_{\!}$ our model design and training targets indeed contribute to our strong results.

\subsection{Performance on Instruction Following}\label{sec:ex_if}
	\vspace*{-2pt}
\noindent\textbf{Dataset.}~We conduct experiments on three VLN datasets:
\begin{itemize}[leftmargin=*]
	\setlength{\itemsep}{0pt}
	\setlength{\parsep}{-2pt}
	\setlength{\parskip}{-0pt}
	\setlength{\leftmargin}{-10pt}
	\vspace{-5pt}
	\item R2R$_{\!}$~\cite{anderson2018vision}:$_{\!}$ It$_{\!}$ has$_{\!}$ four$_{\!}$ splits,$_{\!}$ \ie,$_{\!}$ \texttt{train}$_{\!}$ ($61_{\!}$ scenes,$_{\!}$ $14,039$ instructions), \texttt{val} \texttt{seen} ($61$ scenes, $1,021$ instructions), \texttt{val}  \texttt{unseen} ($11$ scenes, $2,349$ instructions), and \texttt{test} \texttt{unseen}$_{\!}$ ($18_{\!}$ scenes,$_{\!}$ $4,173_{\!}$ instructions).$_{\!}$ There$_{\!}$ are$_{\!}$ no$_{\!}$ overlapping scenes between  \texttt{train} and \texttt{unseen} splits.
	\item R4R$_{\!}$~\cite{jain2019stay}:$_{\!}$ It$_{\!}$ extends$_{\!}$ R2R$_{\!}$ by$_{\!}$ connecting$_{\!}$ two$_{\!}$ close$_{\!}$~tail-to- head$_{\!}$ trajectories$_{\!}$ and$_{\!}$ corresponding$_{\!}$ instructions$_{\!}$ in$_{\!}$ R2R.$_{\!}$ R4R$_{\!}$ contains$_{\!}$ three$_{\!}$ sets,$_{\!}$ \ie,$_{\!}$ \texttt{train}$_{\!}$ ($61$ scenes, $233,613$ instructions), \texttt{val} \texttt{seen} ($61$ scenes, $1,035$
instructions), and \texttt{val} \texttt{unseen} ($11$ scenes, $45,162$ instructions).
    \item REVERIE$_{\!}$~\cite{2020REVERIE}:$_{\!}$ It$_{\!}$ replaces$_{\!}$ detailed$_{\!}$ instructions$_{\!}$~in$_{\!}$~R2R with high-level descriptions of target locations and ob- jects.$_{\!}$ It$_{\!}$ is$_{\!}$ composed$_{\!}$ of$_{\!}$ four$_{\!}$ sets,$_{\!}$ \ie,$_{\!}$ \texttt{train}$_{\!}$ ($53_{\!}$ scenes, $10,466$ instructions), \texttt{val} \texttt{seen} ($61$ scenes, $1,371$ in- structions),$_{\!}$ \texttt{val}$_{\!}$  \texttt{unseen}$_{\!}$ ($10_{\!}$ scenes,$_{\!}$ $3,753_{\!}$ instructions), and \texttt{test} \texttt{unseen} ($16$ scenes, $6,292$ instructions).
	\vspace{-3pt}
\end{itemize}

\noindent\textbf{Evaluation$_{\!}$ Metric.$_{\!}$} For$_{\!}$ R2R,$_{\!}$ we$_{\!}$ follow$_{\!}$ conventions$_{\!}$~\cite{anderson2018vision,fried2018speaker} to~report four evaluation metrics: i) \textit{Success Rate} (\texttt{SR}), ii) \textit{Trajectory Length} (\texttt{TL}), iii) \textit{Oracle success Rate} (\texttt{OR}), and iv) \textit{Success rate weighted by Path Length} (\texttt{SPL}), where$_{\!}$ \texttt{SR}$_{\!}$ and$_{\!}$ \texttt{SPL}$_{\!}$ are$_{\!}$ of$_{\!}$ priority.$_{\!}$ For R4R, we further adopt v) \textit{Coverage weighted by Length Score} (\texttt{CLS})~\cite{jain2019stay}, vi) \textit{normalized Dynamic Time Warping} (\texttt{nDTW}), and vii) \textit{Success weighted by nDTW} (\texttt{SDTW}). For$_{\!}$ REVERIE,$_{\!}$ the$_{\!}$ first four
metrics are also employed for its navigation sub-task, and~viii) \textit{Remote Grounding Success rate} (\texttt{RGS}) and  ix) \textit{RGS weighted by Path Length} (\texttt{RGSPL}) are additionally used for overall performance evaluation.

\noindent\textbf{Quantitative Result.} Several famous and recent advanced solutions$_{\!}$~\cite{fried2018speaker,tan2019learning,zhu2019vision,wang2019reinforced,wang2020active,hao2020towards,hong2020recurrent,deng2020evolving,guhur2021airbert,chen2021history,wang2022counterfactual,qiao2022hop} for instruction following are involved in comparison.
Note that {we report the score of the single model under the \textbf{single run} setup following the tradition~\cite{hong2020recurrent,wang2022counterfactual,qiao2022hop,chen2021history}.} As shown in Table~\ref{table:IFr2r}, \textsc{Lana}$_{st}$, which is only fine-tuned on wayfinding after multi-task pretraining, demonstrates comparable, if not better, results than those alternatives on R2R. Remarkably, {\textsc{Lana}$_{mt}$}, which learns to interpret navigation paths alongside following instructions, even yields better navigation performance. For instance, {\textsc{Lana}$_{mt}$} lifts {\textsc{Lana}$_{st}$} by 2\% and 1\% SPL, on \texttt{val} \texttt{unseen} and \texttt{test} respectively. This verifies the efficacy of our language-capable navigation scheme and multi-task learning strategy. More significant improvements can be observed on R4R (\textit{cf}.$_{\!}$~Table$_{\!}$~\ref{table:IFr4r}) and~REVERIE (\textit{cf}.$_{\!}$~Table$_{\!}$~\ref{table:IFrev}), where the former focuses on long-horizon navigation with longer instructions and trajectories, while the latter gives abstract instructions only.$_{\!}$ These$_{\!}$ results$_{\!}$ confirm$_{\!}$ our$_{\!}$ generality$_{\!}$ and$_{\!}$ versatility.$_{\!}$ It$_{\!}$~is important to note that all the competitors are only aware of wayfinding, while our agent can generate grounded route descriptions for interpreting its navigation behaviors/plans.

\subsection{Performance on Instruction Generation}\label{sec:ex_ig}

\noindent\textbf{Dataset.}~We compare machine generated route descriptions with the human-written instructions, on two VLN datasets:
\begin{itemize}[leftmargin=*]
	\setlength{\itemsep}{0pt}
	\setlength{\parsep}{-2pt}
	\setlength{\parskip}{-0pt}
	\setlength{\leftmargin}{-10pt}
	\vspace{-5pt}
	\item R2R$_{\!}$~\cite{anderson2018vision}:$_{\!}$ As$_{\!}$  R2R$_{\!}$ \texttt{test}$_{\!}$ is$_{\!}$ preserved$_{\!}$ for$_{\!}$ benchmarking$_{\!}$ instruction following agents, we report the performance of instruction generation on \texttt{val}  sets. Each R2R navigation path is associated with three ground-truth instructions.
	\item R4R$_{\!}$~\cite{jain2019stay}:$_{\!}$ Performance$_{\!}$ is$_{\!}$ reported$_{\!}$ on$_{\!}$ R4R$_{\!}$ \texttt{val}$_{\!}$ sets,$_{\!}$ where each path corresponds to nine ground-truth instructions.
	\vspace{-3pt}
\end{itemize}
REVERIE$_{\!}$~\cite{2020REVERIE}$_{\!}$ is$_{\!}$ not$_{\!}$ involved$_{\!}$ as$_{\!}$ its$_{\!}$ instructions$_{\!}$ are$_{\!}$ high-level, concise descriptions of remote objects, which cannot serve our purpose of grounded instruction generation.

\begin{figure*}[t]
	\vspace{-9pt}
	\begin{center}
		\includegraphics[width=0.93\linewidth]{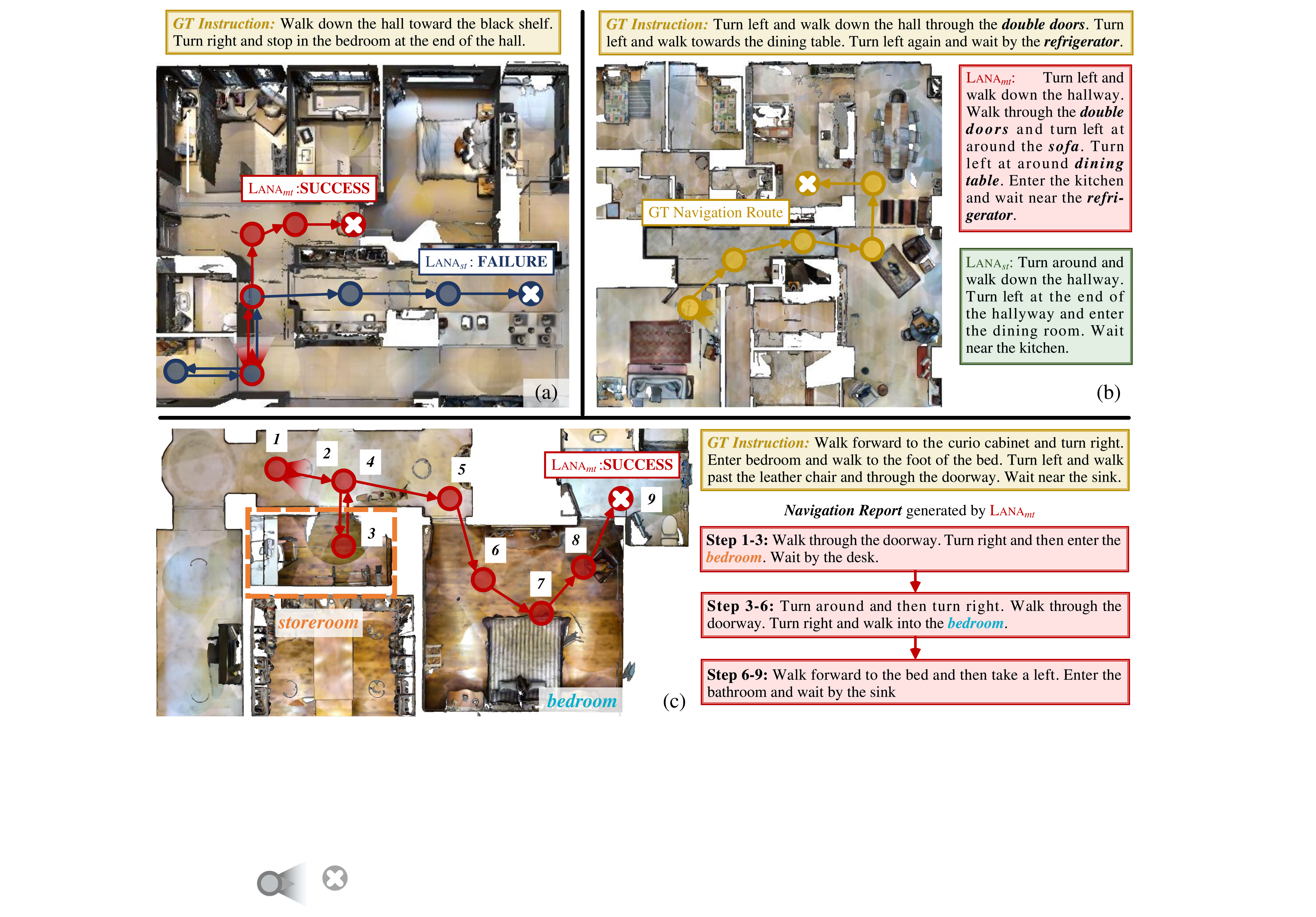}
	\end{center}
	\vspace{-18pt}
	\captionsetup{font=small}
	\caption{\small{(a-b) Visual comparison results between \textsc{Lana}$_{mt}$ and \textsc{Lana}$_{st}$ for (a) instruction following and (b) instruction generation tasks. The start and end points of a navigation route are respectively denoted by \protect\includegraphics[scale=0.07,valign=c]{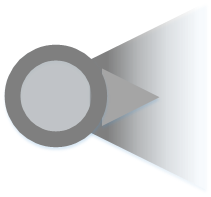} and \protect\includegraphics[scale=0.09,valign=c]{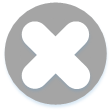}.  (c) \textsc{Lana} is able to interpret its navigation behavior using natural language. The generated report keeps the monitor updated on the navigation process, and even helps reveal the failure mode. For example, at step 2-3, \textsc{Lana}$_{mt}$ enters the storeroom because \textsc{Lana}$_{mt}$ thought it is a bedroom. See \S\ref{sec:ve} for more detailed discussion.}}
	\vspace{-12pt}
	\label{fig:result}
\end{figure*}

\noindent\textbf{Evaluation$_{\!}$ Metric.$_{\!}$} Following$_{\!}$~\cite{agarwal2019visual,wang2022counterfactual},$_{\!}$ we$_{\!}$ opt$_{\!}$ for$_{\!}$ five$_{\!}$ text$_{\!}$ metrics:$_{\!}$ i)$_{\!}$ \texttt{BLEU}$_{\!}$~\cite{papineni2002bleu},$_{\!}$ ii)$_{\!}$ \texttt{CIDEr}$_{\!}$~\cite{vedantam2015cider},$_{\!}$ iii)$_{\!}$ \texttt{METEOR}$_{\!}$~\cite{banerjee2005meteor},$_{\!}$ iv) \texttt{ROUGE}$_{\!}$~\cite{lin2004rouge},$_{\!}$ and$_{\!}$ v)$_{\!}$ \texttt{SPICE}$_{\!}$~\cite{anderson2016spice}.$_{\!}$ For$_{\!}$ each$_{\!}$ navigation$_{\!}$ path,$_{\!}$ the$_{\!}$ metrics are averaged over all the corresponding groundtruth instructions. \texttt{SPICE} is considered as the primary metric.

\noindent\textbf{Quantitative Result.} We compare \textsc{Lana} with four instruction generation algorithms$_{\!}$~\cite{fried2018speaker,tan2019learning,agarwal2019visual,wang2022counterfactual}. Table~\ref{table:R2Rig} and Table~\ref{table:R4Rig} summarize our comparison results. We can find that our$_{\!}$ task-specific$_{\!}$ agent,$_{\!}$ \ie,$_{\!}$ {\textsc{Lana}$_{st}$},$_{\!}$ already$_{\!}$ outperforms$_{\!}$ all the$_{\!}$ competitors,$_{\!}$ across$_{\!}$ all$_{\!}$ the$_{\!}$ metrics$_{\!}$ and$_{\!}$ datasets.$_{\!}$ Note$_{\!}$ that, CCC$_{\!}$~\cite{wang2022counterfactual}, a current top-leading solution, learns the instruction generation model with the aid of a separate wayfinder. More impressively, our multi-task agent, \ie, {\textsc{Lana}$_{mt}$}, performs on par or even better than {\textsc{Lana}$_{st}$}, demonstrating the algorithmic and functional advantages of our approach.

\noindent\textbf{User$_{\!}$ Study.$_{\!}$} To$_{\!}$ provide$_{\!}$ a$_{\!}$ complete$_{\!}$ measure$_{\!}$ of$_{\!}$ the$_{\!}$ quality$_{\!}$~of our created instructions, we conduct a set of human evalua- tion$_{\!}$ experiments,$_{\!}$ based$_{\!}$ on$_{\!}$ pair-wise$_{\!}$ comparison.$_{\!}$ Concretely,$_{\!}$ 50$_{\!}$ college$_{\!}$ students$_{\!}$ are$_{\!}$ asked$_{\!}$ to$_{\!}$ respectively$_{\!}$ compare$_{\!}$ the$_{\!}$ ins- tructions$_{\!}$ generated$_{\!}$ by$_{\!}$ {\textsc{Lana}$_{mt\!}$} with$_{\!}$ those$_{\!}$ created$_{\!}$ by CCC, BT-Speaker, and humans, for 100 paths in total. The paths are
 sampled$_{\!}$ from$_{\!}$ R2R$_{\!}$ \texttt{val}$_{\!}$ \texttt{unseen}.$_{\!}$ Finally,$_{\!}$ {\textsc{Lana}} receive more preference votes, \ie, \textbf{63.4}\% \textit{vs} CCC 36.6\%, and$_{\!}$
 \textbf{75.1}\%$_{\!}$ \textit{vs}$_{\!}$ BT-Speaker$_{\!}$ 24.9\%.$_{\!}$ Yet,$_{\!}$ human-written$_{\!}$ instructions$_{\!}$ are$_{\!}$ far$_{\!}$ more$_{\!}$ favorable,$_{\!}$ \ie,$_{\!}$ {69.3}\%$_{\!}$ \textit{vs}$_{\!}$ {\textsc{Lana}}$_{\!}$ 30.7\%, demonstrating there remains large room for improvement.

\vspace*{-2pt}
\subsection{Diagnostic Experiment}\label{sec:abl}
\vspace*{-2pt}
To thoroughly study the effectiveness of our language-capable navigation framework, we carry out a series of diagnostic experiments on  \texttt{val} \texttt{unseen} set of R2R~\cite{anderson2018vision}, for both instruction following and generation tasks. The experimental results are summarized in Table~\ref{table:R2Rab}.  More specifically, a total of eight baselines are involved in our ablation study:
\begin{enumerate}[leftmargin=*]
	\setlength{\itemsep}{0pt}
	\setlength{\parsep}{-2pt}
	\setlength{\parskip}{-0pt}
	\setlength{\leftmargin}{-10pt}
	\vspace{-5pt}
	\item fine-tune on instruction following only, \textit{w/o} pretraining;
	\item fine-tune$_{\!}$ on$_{\!}$ instruction$_{\!}$ generation$_{\!}$ only,$_{\!}$ \textit{w/o}$_{\!}$  pre-training;$_{\!\!}$
    \item fine-tune on both instruction following and generation, \textit{w/o} pretraining;
    \item pretrain and fine-tune on instruction following only;
	\item pretrain and fine-tune on instruction generation only;
    \item pretrain$_{\!}$ on$_{\!}$ both$_{\!}$ instruction$_{\!}$ following$_{\!}$ and$_{\!}$ generation,$_{\!}$ and fine-tune on instruction following only;
    \item pretrain$_{\!}$ on$_{\!}$ both$_{\!}$ instruction$_{\!}$ following$_{\!}$ and$_{\!}$ generation,$_{\!}$ and fine-tune on instruction generation only;
    \item pretrain and fine-tune on both instruction following and generation.
	\vspace{-3pt}
\end{enumerate}
These baselines can be roughly grouped into three classes: \textbf{i)} baselines 1,2, and 3 are all \textit{w/o} pretraining, and fine-tune on each task either individually or jointly; \textbf{ii)} baselines 4 and 5 pretrain and fine-tune on each task individually; and \textbf{iii)} baselines 6, 7, and 8 are \textit{w/} joint-task pretraining, and fine-tune on each task either individually or jointly. Baselines 6 and 7 are the two sing-task agents, \ie,  \textsc{Lana}$_{st}$, and baseline 8 are our finally delivered agent \textsc{Lana}$_{mt}$; their performance have been thoroughly reported in \S\ref{sec:ex_if} and \S\ref{sec:ex_ig}.

Also, note that baselines 1, 4, and 6 only master wayfinding, while baselines 2, 5, and  7 can only undertake the route description task. Baselines 3 and 8 are capable of both.

Several essential conclusions can be drawn:
\begin{itemize}[leftmargin=*]
	\setlength{\itemsep}{0pt}
	\setlength{\parsep}{-2pt}
	\setlength{\parskip}{-0pt}
	\setlength{\leftmargin}{-10pt}
	\vspace{-5pt}
	\item Joint-task fine-tuning can benefit the performance of both tasks (baseline 3 \textit{vs} 1 \textit{vs} 2);

\item Joint-task pretraining and fine-tuning can benefit the performance of both tasks (baseline 8 \textit{vs} 6 \textit{vs} 7);

	\item Pretraining can facilitate the final performance (baseline 4 \textit{vs} 1, baseline 5 \textit{vs} 2, baseline 6 \textit{vs} 1, baseline 7 \textit{vs} 2, and baseline 8 \textit{vs} 3);

\item Joint-task pretraining is more favored than single-task pretraining (baseline 8 \textit{vs} 4 \textit{vs} 5);

\item Joint-task pretraining and fine-tuning is more favored than all the other training strategies (baseline 8 \textit{vs} 1-7).
	\vspace{-18pt}
\end{itemize}
Note that, joint-tasking pretraining and fine-tuning not only promotes the performance, but increases parameter efficiency, \ie, baseline 8 (143 M) \textit{vs} 6 + 7 (220 M = 123 M + 97 M). In a nutshell, our ablative experiments solidly verify the power of our idea, the efficacy of our algorithmic design, and our advantage in efficient-parameter utilization.

\vspace*{-2pt}
\subsection{Qualitative Experiment}\label{sec:ve}
\vspace*{-3pt}
Fig.~\ref{fig:result} depicts three exemplar navigation episodes from \texttt{val} \texttt{unseen} set of R2R$_{\!}$~\cite{anderson2018vision} . Fig.$_{\!}$~\ref{fig:result}$_{\!}$~(a) compares  \textsc{Lana}$_{mt}$  against  \textsc{Lana}$_{st}$ on the instruction following task. As seen, \textsc{Lana}$_{mt}$ performs robust in this challenging case, while \textsc{Lana}$_{st}$ fails to reach the target location.  As both \textsc{Lana}$_{mt}$  and  \textsc{Lana}$_{st}$ are built with similar network architectures and pretraining protocol, we attribute this to the exploration of cross-task knowledge$_{\!}$ during$_{\!}$ fine-tuning.$_{\!}$ Fig.$_{\!}$~\ref{fig:result}$_{\!}$~(b)$_{\!}$ visua- lizes comparison on instruction creation. As seen, \textsc{Lana}$_{mt}$ outputs more grounded instructions that contain precise action descriptions (\eg, turn left, walk down) as well as sa- lient landmarks (\eg, double doors, dining table, refrigerator). These descriptions have similar properties as human-
generated$_{\!}$ texts,$_{\!}$ even$_{\!}$ involving$_{\!}$ some$_{\!}$ landmarks$_{\!}$ (\eg,$_{\!}$ sofa) that$_{\!}$ are$_{\!}$ informative$_{\!}$ yet$_{\!}$ missed$_{\!}$ in$_{\!}$ human$_{\!}$ reference.$_{\!}$  Fig.$_{\!}$~\ref{fig:result}$_{\!}$~(c) shows that, \textsc{Lana} can offer real-time behavioral interpretation by showing human text report of its navigation process. This not only eases human from consistent monitoring, but also reveals its inner mode to some extent. For example, the report at step 1-3 informs that \textsc{Lana} wrongly recognizes the storeroom as the bedroom -- this is why \textsc{Lana} chooses to enter the storeroom at step 2. In short, as a language-capable navigator, \textsc{Lana} shows advantages in (post-hoc) interpretability and human-robot bi-directional
communication, which are the basic premises of human trust generated.

   \vspace*{-5pt}
\section{Conclusion and Discussion}
   \vspace*{-3pt}
This work calls for a paradigm shift from current VLN
 agents$_{\!}$ --$_{\!}$ strong$_{\!}$ language-aided$_{\!}$ wayfinders$_{\!}$ but$_{\!}$ without$_{\!}$ lan- guage generation ability -- towards more language-capable navigation robots that can not only execute navigation ins- tructions but also verbally describe the navigation routes. We
 present \textsc{Lana}, which learns to master both instruction following and$_{\!}$ generation$_{\!}$ with$_{\!}$ one$_{\!}$ single$_{\!}$ model.$_{\!}$ \textsc{Lana}$_{\!}$ performs on par or even better than previous task-specific solutions in both tasks, with much reduced complexity. Crucially, \textsc{Lana} can write high-quality route descriptions that are informative to interpret its behavior and direct humans in$_{\!}$ collaboration.$_{\!}$ We$_{\!}$ believe$_{\!}$ \textsc{Lana}$_{\!}$ provides$_{\!}$ a$_{\!}$ solid$_{\!}$
basis$_{\!}$~for the$_{\!}$  creation$_{\!}$ of$_{\!}$ language-capable$_{\!}$ robots$_{\!}$ and$_{\!}$ brings$_{\!}$ us$_{\!}$ clo- ser to the ultimate goal of building socially-intelligent and trustworthy$_{\!}$ robots.$_{\!}$ Future$_{\!}$ work$_{\!}$ should$_{\!}$ reinforce$_{\!}$ \textsc{Lana}$_{\!}$ with
the knowledge of large-scale pretrained foundation models.

\newpage


\makesupptitle{\textsc{Lana}: A Language-Capable Navigator for Instruction Following and Generation$_{\!\!\!\!\!\!}$ \\\textit{Supplementary Material}}


This document provides more details of our approach and additional experimental results, organized as follows:
\begin{itemize}
	\vspace{-5pt}
	\setlength{\itemsep}{0pt}
	\setlength{\parsep}{0pt}
	\setlength{\parskip}{0pt}
	\item \S\ref{sec:s1} Implementation Details of \textsc{Lana}.
	\item \S\ref{sec:s2} Additional Quantitative Results on REVERIE$_{\!}$~\cite{2020REVERIE}.
	\item \S\ref{sec:s3} More Ablative Study on Route Encoder.
	\item \S\ref{sec:s4} Additional Qualitative Results.
	\item \S\ref{sec:discuss} Discussion about Social Impact and Limitations.
\end{itemize}

\section{{Implementation Details} of \textsc{Lana}}~\label{sec:s1}
We employ an additional Instruction Trajectory Matching (ITM) task following previous efforts~\cite{chen2021history} during pretraining, which predicts whether a pair of instruction and trajectory is aligned.
The three tasks IF (Instruction Following), IG (Instruction Generation) and ITM (Instruction Trajectory Matching) are sampled with a ratio IG:IF:ITM=4:1:2. We present the pseudo-code of the pretraining procedure in Algorithm~\ref{alg:lana} (ITM is omitted for simplicity).
For finetuning, the instruction following task is optimized with Reinforcement Learning (RL) and Imitation Learning (IL).
IL utilizes the same loss in Eq.\textcolor{red}{13} while RL is implemented based on the Asynchronous Advantage Actor-Critic (A3C) algorithm~\cite{mnih2016asynchronous}. During
finetuning, the sampling ratio for IG and IF is set to IG:IF=2:5; the ITM task is abandoned.
Following the common practice~\cite{chen2021history,hong2020recurrent,qiao2022hop}, we concatenate the object features with the panoramic features and add an object grounding loss for the instruction following task on REVERIE~\cite{2020REVERIE}.
The detailed architecture of \textsc{Lana} is shown in Table~\ref{table:arch}.

\begin{algorithm}
    \caption{The pseudo-code of pre-training for \textsc{Lana}.}\label{alg:lana}
    \textbf{Arguments:} The labeled dataset $\mathcal{H}\!=\!\{(R,X)\}$, the maximum iteration $N$, Route Encoder $\mathcal{E}^{r\!}$, Language$_{\!}$ Encoder$_{\!}$ $\mathcal{E}^{l\!}$, Language$_{\!}$ Decoder$_{\!}$, $\mathcal{D}^{l\!}$, and$_{\!}$ Route$_{\!}$ Decoder$_{\!}$ $\mathcal{D}^{r\!}$.
    \begin{algorithmic}[1]
    \State Initialize $\mathcal{E}^{r\!}$, $\mathcal{E}^{l\!}$, $\mathcal{D}^{r\!}$, $\mathcal{D}^{l\!}$
    \For{iteration $i \in [1,\dots,N]$}
		\State Sample batch $B\subset \mathcal{H}$
		\State Sample a pretraining task $\mathcal{T}$ from \{\textit{IG}, \textit{IF}\}
		\State $\mathcal{L}\gets 0$
		\If{$\mathcal{T}$ is \textit{IG}}
			\For{$(R,X) \in B$}
				\State $[\bar{\bm{r}}_{1:T}]= \mathcal{E}^{r\!}(R)$
				\State $[\bar{\bm{x}}_{1:l-1}]=\!\mathcal{E}^{l}([x_{1:l-1}])$
				\State $\bm{q}_{l}=\mathcal{D}^{l\!}([\bar{\bm{x}}_{1:l-1}], [\bar{\bm{r}}_{1:T}])$
				\State Estimate $\mathcal{L}^g$ \Comment{Defined in Eq.\textcolor{red}{12}.}
				\State $\mathcal{L}\gets \mathcal{L} + \mathcal{L}^g$
			\EndFor
			\State Calculate $\partial\mathcal{L}$
			\State Update $\mathcal{E}^{r\!}$, $\mathcal{E}^{l\!}$, $\mathcal{D}^{l\!}$
		\ElsIf{$\mathcal{T}$ is \textit{IF}}
			\For{$(R,X) \in B$}
				\State $[\bar{\bm{r}}_{1:t-1}, \bar{\bm{O}}_{t}]=\!\mathcal{E}^{r\!}([\bm{r}_{1:t-1}, \bm{O}_{t}])$
				\State $[\bar{\bm{x}}_{1:L}]=\!\mathcal{E}^{l}(X)$
				\State $\bm{p}_{t}=\mathcal{D}^{r\!}([\bar{\bm{r}}_{1:t-1}, \bar{\bm{O}}_{t}], [\bar{\bm{x}}_{1:L}])$
				\State Estimate $\mathcal{L}^f$ \Comment{Defined in Eq.\textcolor{red}{13}.}
				\State $\mathcal{L}\gets \mathcal{L} + \mathcal{L}^f$
			\EndFor
			\State Calculate $\partial\mathcal{L}$
			\State Update $\mathcal{E}^{r\!}$, $\mathcal{E}^{l\!}$, $\mathcal{D}^{r\!}$
		\EndIf
    \EndFor
    \Return $\mathcal{E}^{r\!}$, $\mathcal{E}^{l\!}$, $\mathcal{D}^{r\!}$, $\mathcal{D}^{l\!}$
    \end{algorithmic}
\end{algorithm}

\begin{table}[ht]
\small
\centering
\setlength\tabcolsep{2pt}
\resizebox{\columnwidth}{!}{
\begin{tabular}{c|c|c|c|c}
\hline
 & Language Encoder $\mathcal{E}^{l\!}$ & Route Encoder $\mathcal{E}^{r\!}$ & Language Decoder $\mathcal{D}^{l\!}$ & Route Decoder $\mathcal{D}^{r\!}$ \\
\hline
\multirow{3}{*}{Layer} & \multirow{3}{*}{\(\left[\begin{array}{c}\text{self\_att}\\[-.1em] \text{feedforward}\\[-.1em] \text{-}\end{array}\right]\)$\times$9} & \multirow{3}{*}{\(\left[\begin{array}{c}\text{cross\_att}\\[-.1em] \text{self\_att}\\[-.1em] \text{feedforward}\end{array}\right]\)$\times$1} & \multirow{3}{*}{\(\left[\begin{array}{c}\text{cross\_att}\\[-.1em] \text{self\_att}\\[-.1em] \text{feedforward}\end{array}\right]\)$\times$4}  &  \multirow{3}{*}{\(\left[\begin{array}{c}\text{cross\_att}\\[-.1em] \text{self\_att}\\[-.1em] \text{feedforward}\end{array}\right]\)$\times$4}  \\
 & & & &  \\
 & &  &  &  \\ \hline
\end{tabular}}
\captionsetup{font=small}
		\caption{\small{Detailed model architecture of \textsc{Lana} (\S\ref{sec:s1}).  }} 
\label{table:arch}
\end{table}

\section{$_{\!}$Additional$_{\!}$  Quantitative$_{\!}$  Results$_{\!}$  on$_{\!}$  REVERIE$_{\!\!\!\!\!}$}\label{sec:s2}
The synthetic samples in the PREVALENT dataset are created with a speaker trained on R2R~\cite{anderson2018vision}.
A recent work DUET~\cite{chen2022think} collected a new augmented dataset by synthesizing instructions with a speaker model trained
on the REVERIE dataset~\cite{2020REVERIE}.
We report additional quantitative results of \textsc{Lana} trained with this dataset in Table~\ref{table:IFrevnew}.
Remarkably, this training strategy boosts the performance by a large margin on REVERIE~\cite{2020REVERIE}.
\textsc{Lana} achieves better navigation performance than DUET~\cite{chen2022think} with the same training set, demonstrating the algorithmic advantages of our approach.

\begin{table*}[t]
	\centering
	\resizebox{0.99\textwidth}{!}{
	\setlength\tabcolsep{7pt}
	\renewcommand\arraystretch{1.03}
	\begin{tabular}{|rl||cccccc|cccccc|}
	\hline \thickhline
	   ~ & & \multicolumn{6}{c|}{REVERIE \texttt{val} \texttt{unseen}} & \multicolumn{6}{c|}{REVERIE \texttt{test} \texttt{unseen}} \\
	  \cline{3-14}\cline{3-14}\cline{3-14}\cline{3-14}
	\multicolumn{2}{|c||}{\multirow{-2}{*}{Methods}}
	&\textbf{\texttt{SR}}$_{\!}$~$\uparrow$  &\textbf{\texttt{SPL}}$_{\!}$~$\uparrow$ &\texttt{OR}~$\uparrow$  &\texttt{TL}~$\downarrow$ &\texttt{RGS}$_{\!}$~$\uparrow$&\texttt{RGSPL}$_{\!}$~$\uparrow$
	&\textbf{\texttt{SR}}$_{\!}$~$\uparrow$ &\textbf{\texttt{SPL}}$_{\!}$~$\uparrow$ &\texttt{OR}$_{\!}$~$\uparrow$  &\texttt{TL}~$\downarrow$ &\texttt{RGS}$_{\!}$~$\uparrow$&\texttt{RGSPL}$_{\!}$~$\uparrow$\\
	\hline
	\hline
	RCM~\cite{wang2019reinforced}\!\!&\!\!\!\!\pub{CVPR2019}  &9.29&6.97&14.23&11.98&4.89&3.89&7.84&6.67&11.68&10.60&3.67&3.14\\
	VLN$\circlearrowright$BERT~\cite{hong2020recurrent}\!\!&\!\!\!\!\pub{CVPR2021}&30.67&24.90&35.02&16.78&18.77&15.27&29.61&23.99&32.91&15.86&16.50&13.51\\
	AirBERT~\cite{guhur2021airbert}\!\!&\!\!\!\!\pub{ICCV2021}&27.89&21.88&34.51&18.71&18.23&14.18&30.28&23.61&34.20&17.91&16.83&13.28\\
	HAMT~\cite{chen2021history}\!\!&\!\!\!\!\pub{NeurIPS2021} &32.95 &30.20  &36.84  &14.08 &18.92 &17.28  &30.40 &26.67  &33.41 &13.62 &14.88 &13.08\\
	HOP~\cite{qiao2022hop}\!\!&\!\!\!\!\pub{CVPR2022} & 30.39 & 25.10 & 35.30  & 17.16 & 18.23 &15.31 & 29.12 &23.37 & 32.26  & 17.05 &17.13 &13.90 \\
	DUET$^\dagger$~\cite{chen2022think}\!\!&\!\!\!\!\pub{CVPR2022} & 46.98 & 33.73 & 51.07  & 22.11 & 32.15 &23.03 & 52.51 &36.06 & 56.91  & 21.30 &31.88 &22.06 \\
	\hline
	  \textsc{Lana}\!\!&\!\!\!\!\footnotesize{(\textbf{\texttt{ours}})} & 34.00 & 29.26 &  38.54  & 16.28 & 19.03 & 16.18& 33.50 & 26.89 & 36.41  & 16.75 & 17.53 & 14.25  \\
	  \textsc{Lana}$^\dagger~$\!\!&\!\!\!\!\footnotesize{(\textbf{\texttt{ours}})} & \textbf{48.31} & \textbf{33.86} &  \textbf{52.97}  & 23.18 & \textbf{32.86} & 22.77 & 51.72 & \textbf{36.45} & \textbf{57.20}  & 18.83 & \textbf{32.95} & \textbf{22.85}  \\
	\hline
	\end{tabular}
	}
	\captionsetup{font=small}
	\caption{Additional quantitative results for \textbf{instruction following} on REVERIE$_{\!}$~\cite{2020REVERIE}. $\dagger$ indicates the model is trained on the DUET dataset~\cite{chen2022think}. See \S\ref{sec:s2} for details.}\label{table:IFrevnew}
	\end{table*}

	\begin{table*}[t]
		\centering
				\resizebox{0.99\textwidth}{!}{
				\setlength\tabcolsep{8pt}
				\renewcommand\arraystretch{1.03}
		\begin{tabular}{|l|cc||cccc|cccccc|}
		\hline \thickhline
		\multirow{2}{*}{\#}& \multicolumn{2}{|c||}{Route Encoder} & \multicolumn{4}{c|}{Instruction Following} & \multicolumn{6}{c|}{Instruction Generation} \\
		\cline{2-13}\cline{2-13}\cline{2-13} & \texttt{self\_{att}} & \texttt{cross\_{att}}
		&\textbf{\texttt{SR}}~$\uparrow$ &\textbf{\texttt{SPL}}~$\uparrow$ &\texttt{OR}~$\uparrow$  &\texttt{TL}~$\downarrow$
		&\textbf{\texttt{SPICE}}~$\uparrow$  &\texttt{Bleu-1}~$\uparrow$ &\texttt{Bleu-4}~$\uparrow$ &\texttt{CIDEr}~$\uparrow$ &\texttt{Meteor}~$\uparrow$ &\texttt{Rouge}~$\uparrow$ \\
		\hline
        \hline
		1 & \cmark & & 65.6 & 60.1 & 72.7 & 11.7 & 0.202 & 0.703 & 0.262 & 0.375 & 0.224 & 0.476 \\
		2  & & \cmark & 64.8 & 59.8 & 72.8 & 11.9 & 0.222 & 0.715 & 0.281 & 0.430 & 0.233 & 0.486 \\
		3  & \cmark & \cmark&  \textbf{67.9}  & \textbf{61.6} & \textbf{75.7} & 12.0 &\textbf{0.226} &  \textbf{0.736} &  \textbf{0.298} & \textbf{0.457} &  \textbf{0.238} &  \textbf{0.498} \\
		\hline
		\end{tabular}
		}
			\vspace*{-2pt}
		\captionsetup{font=small}
			\caption{\small{Ablation study on R2R \texttt{val} \texttt{unseen}}~\cite{anderson2018vision}. See \S\ref{sec:s3} for details.}
			\label{table:R2Rabnew}
		\vspace*{-2pt}
	\end{table*}

\section{More Ablative Study on Route Encoder}~\label{sec:s3}
In this section, we further study the efficacy of our route encoder design. Our route encoder $\mathcal{E}^r$ considers both previous action tokens $\{\bm{a}_t\}_t$ as well as past panoramic observations $\{\bm{O}_t\}_t$ (see Eq.~{\color{red}3}). We therefore report two variants, whose route encoder 1) only performs the temporal self-attention over the previous action tokens $\{\bm{a}_t\}_t$, and 2) only adopts the cross-attention operation to perceive the historical panoramic observation $\{\bm{O}_t\}_t$.

The results on R2R \texttt{val} \texttt{unseen}~\cite{anderson2018vision} are summarized in Table~\ref{table:R2Rabnew}. As seen, integrating both the historical panoramic observation and previous action information yields the best performance.

\section{Additional Qualitative Results}~\label{sec:s4}
\begin{figure}[ht]
	\vspace{-9pt}
	\begin{center}
		\includegraphics[width=1.0\linewidth]{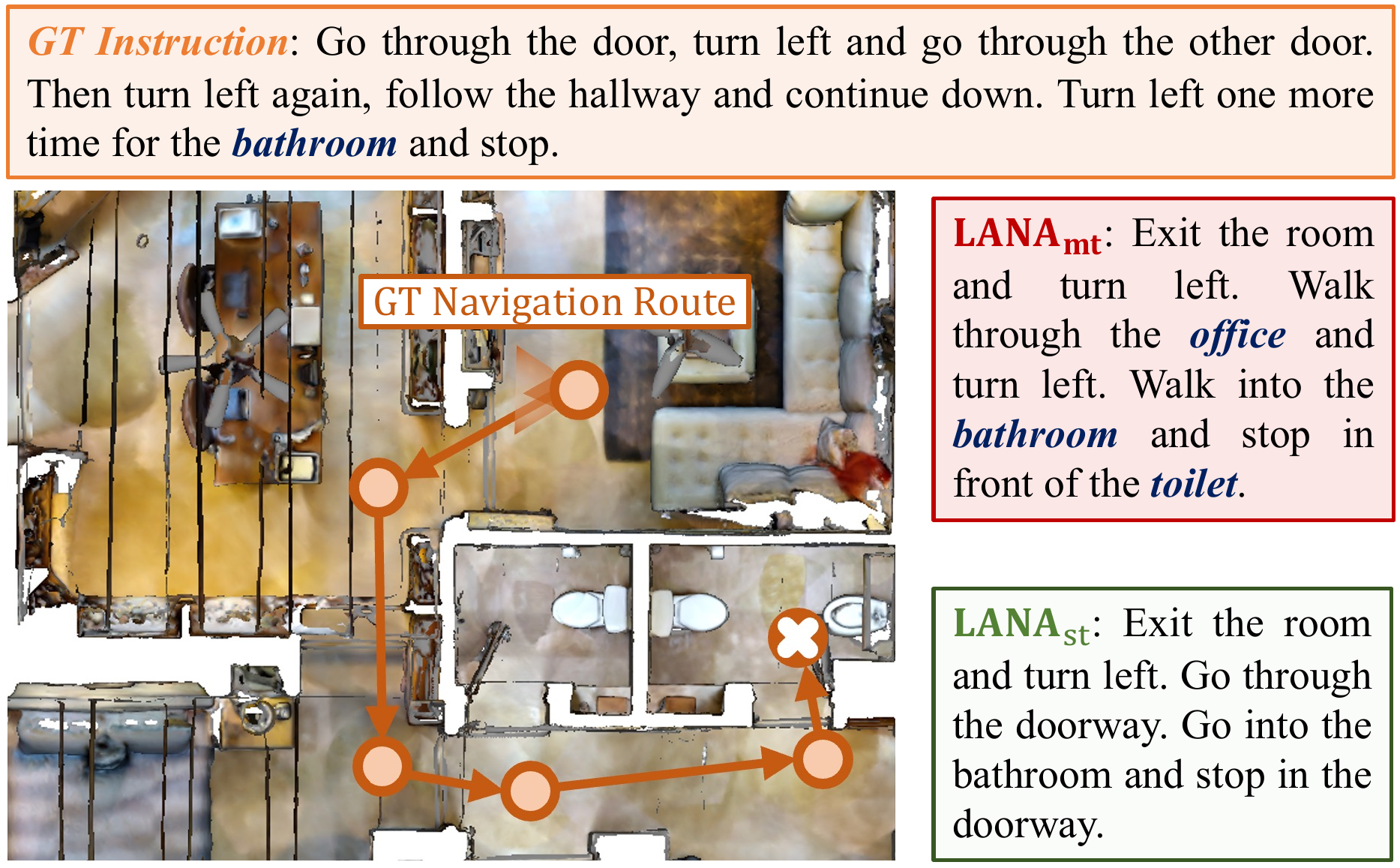}
	\end{center}
	\vspace{-18pt}
	\captionsetup{font=small}
	\caption{$_{\!}$Visual$_{\!}$ comparison$_{\!}$ results$_{\!}$ between$_{\!}$ \textsc{Lana}$_{mt\!}$ and$_{\!}$ \textsc{Lana}$_{st\!}$ for the instruction generation task (\S\ref{sec:s4}). The start and end points of a navigation route are respectively denoted by \protect\includegraphics[scale=0.07,valign=c]{fig/start} and \protect\includegraphics[scale=0.09,valign=c]{fig/end}.}
	\vspace{-12pt}
	\label{fig:result1}
\end{figure}

In this section, we provide more qualitative results for
instruction following and generation. Fig.~\ref{fig:result1} visualizes the comparison between \textsc{Lana}$_{mt}$ and \textsc{Lana}$_{st}$ on instruction generation.
We can observe that \textsc{Lana}$_{mt}$ generates more accurate and vivid instructions. Concretely, \textsc{Lana}$_{mt}$
is able to not only describe precise actions (\eg, turn left, walk through), but also highlight crucial landmark (\eg, office, bathroom, toilet).

\begin{figure*}[ht]
	\begin{center}
		\includegraphics[width=0.8\linewidth]{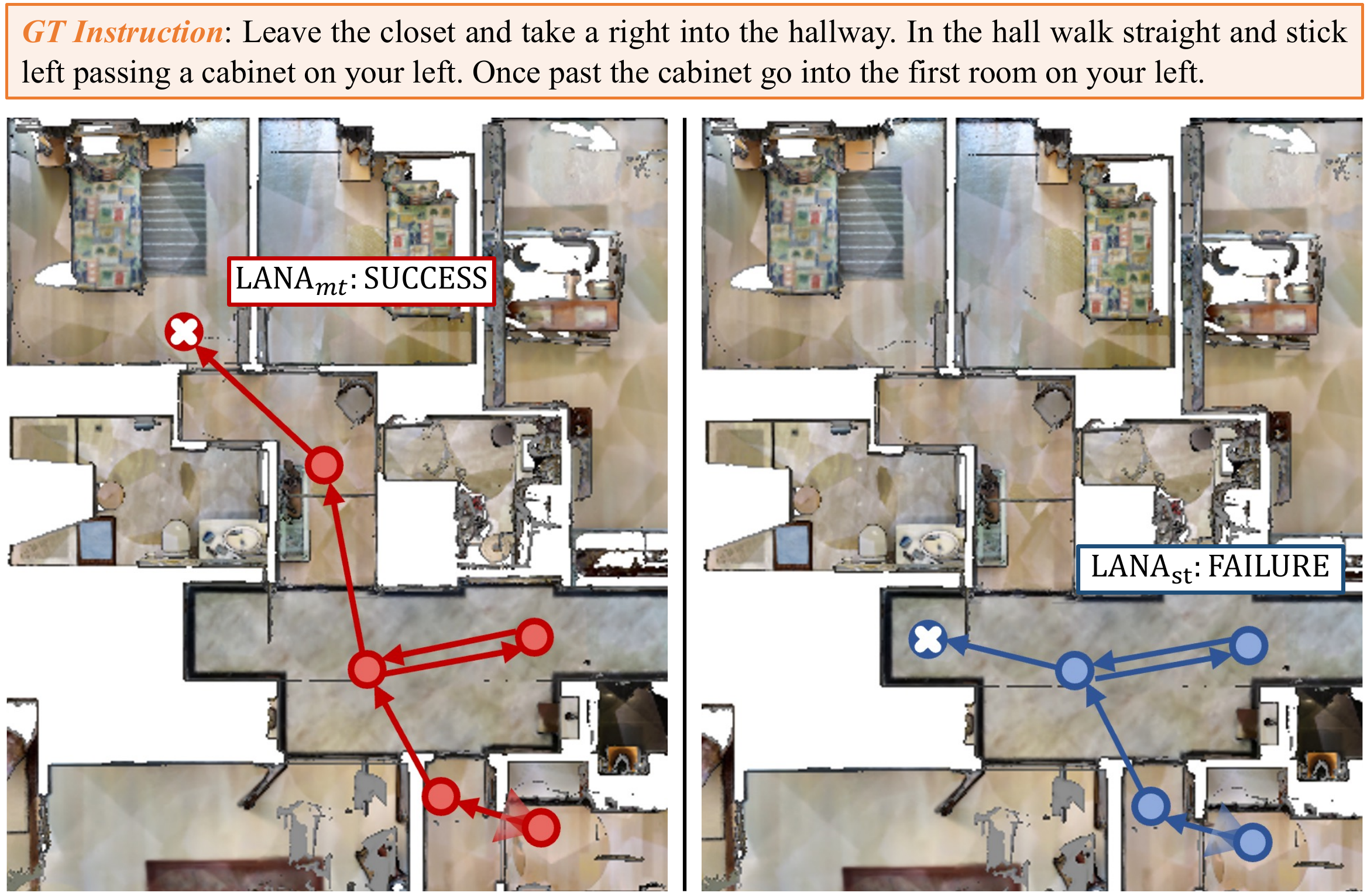}
	\end{center}
	\vspace{-18pt}
	\captionsetup{font=small}
	\caption{Visual comparison results between \textsc{Lana}$_{mt}$ and \textsc{Lana}$_{st}$ for the instruction following task (\S\ref{sec:s4}). The start and end points of a navigation route are respectively denoted by \protect\includegraphics[scale=0.07,valign=c]{fig/start} and \protect\includegraphics[scale=0.09,valign=c]{fig/end}.}
	\vspace{-2pt}
	\label{fig:result2}
\end{figure*}

Fig.~\ref{fig:result2} compares \textsc{Lana}$_{mt}$ with \textsc{Lana}$_{st}$ on instruction following. Given the challenging instruction ``\textit{Leave the closet $\cdots$ on your left}'', \textsc{Lana}$_{mt}$ successfully
take actions to reach the target location, while \textsc{Lana}$_{st}$ terminates the navigation at a wrong position. This intuitively demonstrates the effectiveness of the joint-training strategy.

\begin{figure*}[ht]
	\begin{center}
		\includegraphics[width=0.8\linewidth]{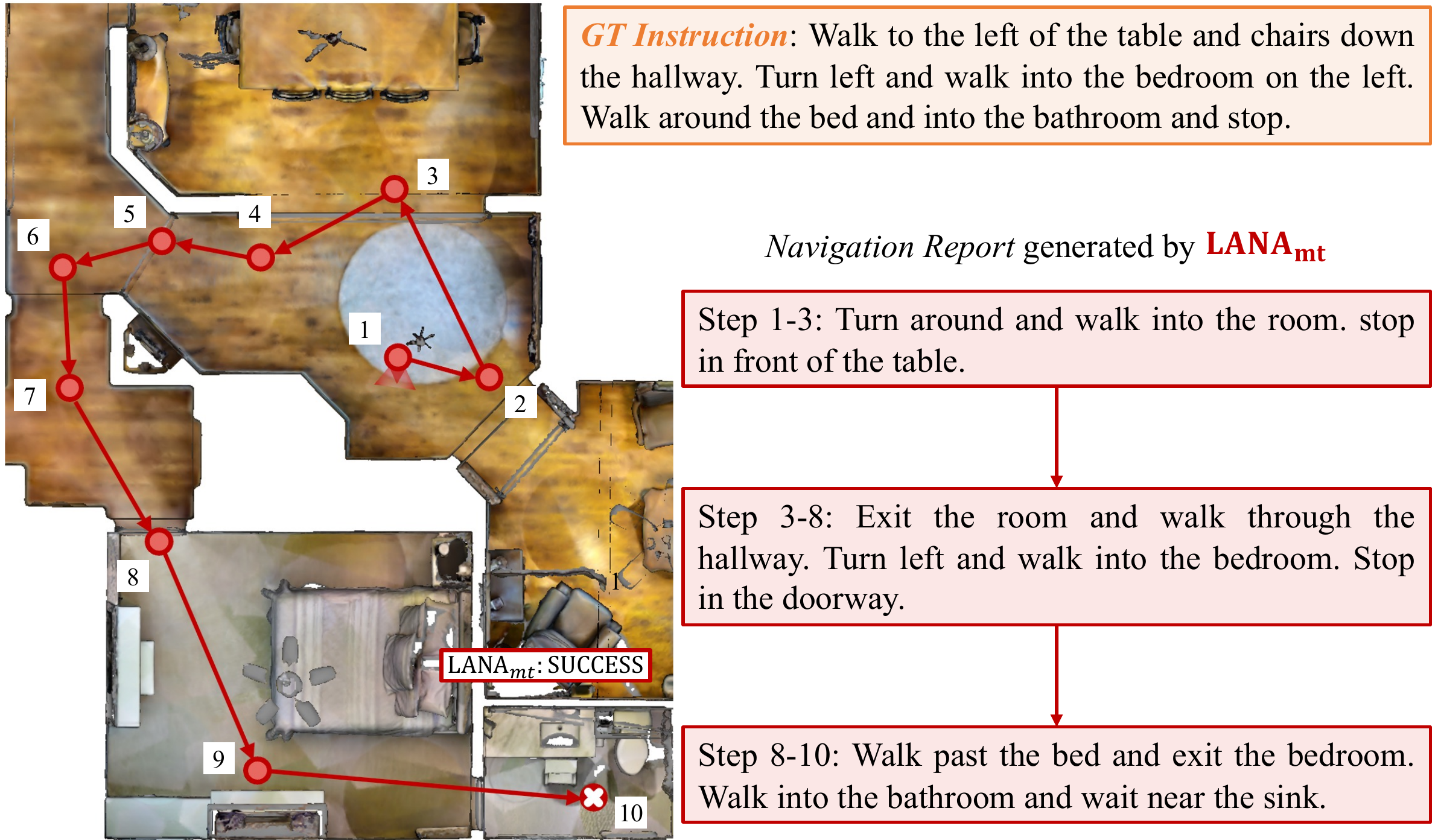}
	\end{center}
	\vspace{-18pt}
	\captionsetup{font=small}
	\caption{Step-by-step navigation behavioral explanation (\S\ref{sec:s4}). The start and end points of a navigation route are respectively denoted by \protect\includegraphics[scale=0.07,valign=c]{fig/start} and \protect\includegraphics[scale=0.09,valign=c]{fig/end}. \textsc{Lana} is able to interpret its navigation behavior using natural language. For example, at step 2-4, \textsc{Lana}$_{mt}$ enters the room and then exits the room because \textsc{Lana}$_{mt}$ intends to find the table mentioned in the instruction.}
	\vspace{-2pt}
	\label{fig:result3}
\end{figure*}

Fig.~\ref{fig:result3} shows the real-time behavioral description provided by \textsc{Lana}$_{mt}$.
The generated report keeps the monitor updated on the navigation process, and reveals its inner decision mode. For examples, the route descriptions generated for Step 1-3 and Step 3-8 can vividly explain to human how \textsc{Lana}$_{mt}$ executes the complex command ``\textit{Walk to the left of the table and chairs down the hallway}'' -- first ``\textit{Walk into the room. Stop in front of the table},'' then ``\textit{exit the room and walk through the hallway}''. This case reveals the advantages of \textsc{Lana} in interpretability and human-robot communication.

\section{Discussion}
\label{sec:discuss}
\noindent\textbf{Social Impact.}
A language-capable navigator can find much broader application scenarios compared with previous ``dumb'' ones and can be more deeply involved into human daily life. It can also serve as a guide robot to assist people who are low-vision or blind.

\noindent\textbf{Limitations.} The agent is developed in virtual simulated environments. If the algorithm is deployed on a real robot in a real dynamic environment, the collisions during navigation can potentially cause damage to persons and assets. More work should be done to practice real-world deployment, \eg, introducing hard constraints to the action space to avoid collisions, and including additional experiments to study the risk of potential damage. In addition, the generated route description, though informative and readable for human, is a kind of post-hoc interpretation. It cannot perfectly and exactly explain the inner decision mode of the agent.

\noindent\textbf{Future work.} In the future, in addition to investigating the efficacy of our approach in other navigation tasks  (\eg, object-goal navigation, audio-goal navigation), we will design more compact architecture to jointly learn the two tasks.

{\small
\bibliographystyle{ieee_fullname}
\bibliography{egbib}
}

\clearpage

\end{document}